%% file: main.tex
\UseRawInputEncoding 

\documentclass[10pt]{article}
\usepackage[letterpaper, left=1in, top=1in, right=1in, bottom=1in, verbose, ignoremp]{geometry}

\usepackage{url}\RequirePackage[colorlinks,citecolor=blue, linkcolor=blue,urlcolor = blue, hyperfootnotes=false]{hyperref}
\usepackage{latexsym,amssymb,amsmath,amsfonts,graphicx,color,fancyvrb,amsthm,enumerate,subcaption,mathrsfs}
\usepackage[,authoryear,round]{natbib} 
\usepackage{pgfplotstable}
\usepackage[dvipsnames]{xcolor}
\usepackage{xy}\xyoption{all} \xyoption{poly} \xyoption{knot}
\usepackage{float}
\restylefloat{table}
\usepackage{bm}
\usepackage{bbm}
\usepackage{multicol,multirow}
\usepackage{array}
\usepackage{relsize}
\usepackage{chngcntr}
\usepackage{etoolbox}
\usepackage{caption}
\usepackage{tikz}
\usetikzlibrary{decorations.pathreplacing,calc}
\usetikzlibrary{shapes,backgrounds}
\usetikzlibrary{patterns}
\usetikzlibrary{cd}
\usepackage{amsopn}
\usepackage{calc}
\usepackage{algorithm}
\usepackage[noend]{algpseudocode}
\usepackage{mathtools}

\usepackage{tikz-cd} 
\usepackage{amscd} 
\usepackage{comment}
\usepackage[capitalize,nameinlink,compress]{cleveref}
\usepackage{pgfplots}
\usepgfplotslibrary{groupplots}
\crefname{figure}{Figure}{Figures} 
\crefname{equation}{}{} 
\crefname{assumption}{Assumption}{Assumptions}
\crefname{subsection}{Subsection}{Subsections}
\usepackage{hypcap}
\usepackage[symbol]{footmisc}
\usepackage{booktabs}
\usepackage{siunitx}
\usepackage{listings}
\definecolor{darkblue}{RGB}{0,0,139}
\definecolor{darkred}{RGB}{180,0,0}
\definecolor{darkgreen}{RGB}{0,120,0}
\definecolor{codebeige}{RGB}{252, 246, 227}
\lstdefinelanguage{Julia}{
  sensitive=true,
  alsoletter={_},
  morekeywords={baremodule,begin,break,catch,const,continue,do,else,elseif,
    end,export,false,finally,for,function,global,if,import,let,local,macro,
    module,quote,return,struct,true,try,using,while,where,mutable,primitive},
  morekeywords=[2]{AffineLayer,ActivationLayer,BigInt,HiGHSMode,
    NeuralNetwork,Rational,Signomial},
  morekeywords=[3]{comp,hoffman_constant,linear_regions,
    lower_hoffman_constant,max,println,prune,relu,tropicalize,
    upper_hoffman_constant},
  morecomment=[l]{\#},
  morecomment=[n]{\#=}{=\#},
  morestring=[b]",
  morestring=[b]'
}
\graphicspath{{./}}

\newcounter{cdrow}

\newtheorem{theorem}{Theorem}[]
\newtheorem*{theorem*}{Theorem}
\newtheorem{corollary}[theorem]{Corollary}
\newtheorem{lemma}[theorem]{Lemma}
\newtheorem{proposition}[theorem]{Proposition}

\newtheorem*{claim*}{Claim}

\theoremstyle{definition}
\newtheorem{definition}[theorem]{Definition}
\newtheorem*{definition*}{Definition}

\theoremstyle{remark}
\newtheorem{remark}[theorem]{Remark}

\newtheorem{example}[theorem]{Example}
\newtheorem*{example*}{Example}
\newtheorem*{notation}{Notation}

\makeatletter
\newcommand*{\op}{%
  \DOTSB
  \mathop{\vphantom{\bigoplus}\mathpalette\matt@op\relax}%
  \slimits@
}
\newcommand\matt@op[2]{%
  \vcenter{\m@th\hbox{\resizebox{\widthof{$#1\bigoplus$}}{!}{$\boxplus$}}}%
}
\makeatother

\DeclareMathOperator{\codim}{\mathrm{codim}}

\newcommand{\R}{\mathbb{R}}

\newcommand{\RR}{{\mathbb{R}}}

\newcommand{\NN}{\mathbb{N}}

\newcommand{\sgn}{\text{\sgn}}

\newcommand{\id}{\text{id}}

\renewcommand{\vec}[1]{\mathbf{#1}}

\def\va{{\vec{a}}}
\def\vb{{\vec{b}}}
\def\vc{{\bm{c}}}
\def\vd{{\bm{d}}}

\def\vf{{\vec{f}}}

\def\vh{{\vec{h}}}

\def\vk{{\bm{k}}}

\def\vn{{\bm{n}}}

\def\vp{{\vec{p}}}
\def\vq{{\vec{q}}}

\def\vu{{\bm{u}}}
\def\vv{{\bm{v}}}

\def\vx{{\vec{x}}}
\def\vy{{\bm{y}}}

\def\mA{{\bm{A}}}
\def\mB{{\bm{B}}}

\def\mF{{\bm{F}}}

\def\mM{{\bm{M}}}
\def\mN{{\bm{N}}}

\def\mW{{\bm{W}}}

\makeatletter
\def\@biblabel#1{}
\makeatother

\makeatletter
\patchcmd{\NAT@citex}
  {\@citea\NAT@hyper@{%
     \NAT@nmfmt{\NAT@nm}%
     \hyper@natlinkbreak{\NAT@aysep\NAT@spacechar}{\@citeb\@extra@b@citeb}%
     \NAT@date}}
  {\@citea\NAT@nmfmt{\NAT@nm}%
   \NAT@aysep\NAT@spacechar\NAT@hyper@{\NAT@date}}{}{}

\patchcmd{\NAT@citex}
  {\@citea\NAT@hyper@{%
     \NAT@nmfmt{\NAT@nm}%
     \hyper@natlinkbreak{\NAT@spacechar\NAT@@open\if*#1*\else#1\NAT@spacechar\fi}%
       {\@citeb\@extra@b@citeb}%
     \NAT@date}}
  {\@citea\NAT@nmfmt{\NAT@nm}%
   \NAT@spacechar\NAT@@open\if*#1*\else#1\NAT@spacechar\fi\NAT@hyper@{\NAT@date}}
  {}{}

\makeatother

\pgfplotsset{compat=1.18} 
\begin{document}

\def\spacingset#1{\renewcommand{\baselinestretch}%
{#1}\small\normalsize} \spacingset{1}

\begin{flushleft}
{\Large{\textbf{A Computational Tropical Geometry Framework\\ for Neural Networks}}}
\newline
\\
Paul Lezeau$^{1,2,\dagger}$, Thomas Walker$^3$, Yueqi Cao$^{4}$, 
Shiv Bhatia$^{1,5}$, and Anthea Monod$^{1}$
\\
\bigskip
\bf{1} Department of Mathematics, Imperial College London, UK
\\
\bf{2} London School of Geometry and Number Theory, UK
\\
\bf{3} Department of Electrical and Computer Engineering, Rice University, USA
\\
\bf{4} Department of Mathematics, KTH Royal Institute of Technology, Sweden
\\
\bf{5} Department of Computing, Imperial College London, UK
\\
\bigskip
$\dagger$ Corresponding e-mail: p.lezeau23@imperial.ac.uk
\end{flushleft}



\section*{Abstract}
We propose a computational tropical geometry framework for the symbolic analysis of neural networks with tropical activations.
The number of linear regions of a neural network has been actively studied as a measure of the expressivity of a given architecture.
To study these, we work in the setting of tropical geometry---a combinatorial and polyhedral variant of algebraic geometry---where there are known connections between tropical rational maps and feedforward neural networks. 
We expand this connection by developing concrete computational tools for studying the linear regions of neural networks.
We present an algorithm, together with a proof of correctness, which computes the linear regions of a neural network as explicit unions of polyhedra. We further relate the computation of the number of linear regions of a tropical expression to the number of monomials that appear in it, and show how tropical expressions can often be pruned to remove redundant monomials.
We introduce the Hoffman constant of a neural network's tropical expression, a geometric quantity that controls the distance from any point in the input space to the farthest linear region. 
We provide the open source Julia library \textsc{TropicalNN.jl}, which is built on top of the OSCAR computer algebra system and implements the algorithms mentioned above to analyze neural networks symbolically using their tropical representations.
We present a set of proof-of-concept computational examples to demonstrate how our tropical geometric theory can be applied to reveal insights on the expressivity of a network architecture.

\input{arXiv/intro_new}

\input{arXiv/background}

\input{arXiv/expressivity}

\input{arXiv/library}

\input{arXiv/experiments}

\input{arXiv/discussion_new}

\section*{AI Usage Disclosure}
The authors used generative AI for the following: supporting language and minor stylistic improvements in the manuscript, writing and debugging Julia code, proofreading, and identifying missing content from the literature review. All AI outputs were generated under careful human guidance, reviewed and verified by the authors, who take full responsibility for the final content.

\section*{Acknowledgments}

We would like to thank the authors of \cite{stargalla2025computational} for noticing an error in a previous version of this manuscript. We would also like to thank Aideen Fay, Kamillo Ferry, Yue Ren, Victoria Schleis, and Roan Talbut for thoughtful feedback and helpful discussions on this work. 

P.L.~was funded by a London School of Geometry and Number Theory--Imperial College London/King's College London/University College London PhD studentship, which is supported by the Engineering and Physical Sciences Research Council [EP/S021590/1]. A.M.~is supported by the EPSRC AI Hub on Mathematical Foundations of Intelligence: An ``Erlangen Programme'' for AI [EP/Y028872/1].


\clearpage
\newpage

\bibliography{arXiv/references}

\end{document}

%% file: arXiv/intro_new.tex
\section{Introduction}

Neural networks are typically represented and analyzed numerically, through their weights, activations, and evaluations on data.  However, networks with piecewise linear activations also possess rich algebraic and geometric structure that can be studied symbolically.  Developing computational tools that expose this structure provides a complementary perspective on neural networks, which then allows properties of the functions they represent to be analyzed using techniques from algebra and geometry.

\emph{Tropical geometry} provides a natural framework for such an approach, as a piecewise linear and polyhedral analogue of algebraic geometry in which addition is replaced by $\max$ and multiplication by ordinary addition.  Objects from tropical geometry therefore combine algebraic representations with explicit polyhedral geometry and admit computational techniques from combinatorics, optimization, and computer algebra \citep[e.g.,][]{mikhalkin2009tropical, speyer2009tropical, maclagan2015introduction}.  Importantly for our setting, feedforward neural networks with ReLU activations admit representations as tropical rational functions \citep{zhang2018tropical}.  This connection makes it possible to treat neural networks as symbolic tropical objects and to use tools from computational tropical and polyhedral geometry for their analysis.

In this work, we develop the tropical geometric perspective into a computational framework for the symbolic analysis of neural networks and demonstrate its use through the study of neural network \emph{expressivity}.  One of the central questions in deep learning is to understand the capacity of neural network architectures to represent complex functions (see \cite{hu2021model} for a survey).  For networks with piecewise linear activation functions, a widely studied measure of this capacity is the number of \emph{linear regions} into which the network partitions its input space \citep[e.g.,][]{pascanu2013number, montufar2014number, arora2016understanding, raghu2017expressive, hanin2019deep, xiong2020number, goujon2024number, montufar2022sharp}.  Each linear region is a maximal domain on which the network computes a single affine map.  While existing work has often focused on counting these regions, the tropical representation makes their underlying polyhedral geometry explicitly accessible.  We exploit this structure both to compute linear regions exactly and to introduce new geometric and algebraic quantities with which to study neural networks.

\paragraph{Contributions.}

In this paper, we establish algebraic and geometric computational tools to study the expressivity of neural networks from a tropical perspective. The contributions of our work are the following:
\begin{itemize}
    \item \textbf{Theoretical:} We establish three theoretical contributions. First, we provide an exact algorithm (\cref{alg:linear-regions}) for computing the linear regions of any rational signomial---and therefore any ReLU neural network---as explicit unions of polyhedra, together with a proof of correctness (\cref{thm:linear-region-algorithm}). We provide an extension of this method in \cref{alg:layerwise-linear-regions}, which computes the linear regions for arbitrary compositions of rational signomials. This provides a unified way of computing the linear regions of neural networks with arbitrary signomial activations.
    To the best of our knowledge, this is the first algorithm that directly leverages the tropical expression of a neural network to obtain such explicit representations of linear regions in this generality. Second, we introduce the \emph{Hoffman constant} of a rational signomial as a geometric quantity that controls the maximal distance from any point in the input space to the farthest linear region. This yields an \emph{effective sampling radius} $R$ such that the ball of radius $R$ is guaranteed to intersect every linear region, turning what is otherwise an enumeration problem into a bounded one. Third, we study the connection between the linear regions of a neural network and the monomials of its tropical expression. The number of non-redundant monomials can be seen as an algebraic measure of the complexity of a tropical representation of a neural network.

    \item \textbf{Experimental:} We provide the new open source library \textsc{TropicalNN.jl}, designed for the exploration of tropical properties of neural networks. This library is built on the Open Source Computer Algebra Research (OSCAR) system \citep{OSCAR}, and can be used to convert both trained and untrained arbitrary neural networks with tropical activations into tropical algebraic objects that can be manipulated symbolically, which then allows, for example, the computation of the exact number of linear regions of the network.
\end{itemize}

The remainder of this paper is organized as follows. We close this section with an overview of related work. \cref{sec:background} then provides background on tropical algebra and its connection to neural networks.
\cref{sec:expressivity} introduces our theoretical contributions: the exact linear region algorithm, the Hoffman constant, and the study of monomials.
\cref{sec:library} presents the \textsc{TropicalNN.jl} library and its key functionalities, together with some computational details.
\cref{sec:experiments} presents our experimental evaluation.

\paragraph{Related Work.} 
Tropical geometry has emerged as a powerful tool for analyzing deep neural networks with piecewise linear activation functions, such as rectified linear units (ReLUs) and maxout units. 
\cite{zhang2018tropical} established that neural networks with ReLU activations can be represented by tropical rational signomials, enabling the use of tropical techniques to study network expressivity. 
They also showed that the decision boundary of a deep ReLU network is contained in a tropical hypersurface. 
This initial work focused on neural networks with Euclidean input domains. \cite{yoshida2024tropical} later extended this connection to incorporate the tropical projective torus as an input domain, broadening the applicability of tropical methods.
Recently, \cite{pasque2026adversarially} leveraged tropical geometry to construct convolutional neural networks with improved robustness against adversarial attacks, and \cite{brandenburg2024realtropicalgeometryneural} leveraged the tropical forms of neural networks to study tropical geometric aspects of classification tasks.

Measures for quantifying neural network expressivity are important for facilitating the theoretical and empirical investigation of neural network properties. Neural networks with piecewise linear activations compute piecewise linear functions on areas of the input space referred to as the network's \emph{linear regions} (as \cite{stargalla2025computational} noted, there exist several concurrent definitions for linear regions used in the literature; see also \cref{rmk:linear-regions-definitions}). The number of distinct linear regions of the network provides an appropriate and quantifiable measure of its expressivity \citep[e.g.,][]{montufar2014number}.
Subsequent research has also worked towards enumerating the number of linear regions a given neural network has. 
Using mixed-integer programming, \cite{serra2018bounding} provide an exact enumeration of the number of activation regions in a bounded subset of the input domain, and
\cite{serra2020empirical} give an empirical upper bound on the maximum number of activation regions of a neural network along with a probabilistic lower bound.
For unbounded domains, \cite{charisopoulos2018tropical} provide a probabilistic method for estimating the linear regions of a one-layer network, together with some analytic bounds for maxout networks. Such networks are also studied by \cite{montufar2022sharp}, which provide asymptotically sharp bounds for the number of linear regions that can arise for these architectures.
Using tropical methods, \cite{piwek2023exact} provide an algorithm for determining the exact number of boundary pieces of a ReLU network, as well as the number of affine pieces of the network. 
In a different direction, \cite{tran2024minimal} initiated the study of the complexity of \emph{tropical representations} of piecewise linear functions beyond counting linear regions, and introduce new invariants such as factorization complexity.

A separate line of research has focused on studying the \emph{geometry} of linear regions from a computational perspective. \cite{zhang2019empirical} take an empirical approach and provide a method to compute the activation region containing a given point in the input space. \cite{wang2022estimation} computes linear regions as unions of activation regions, and \cite{masden2025algorithmic} presents an approach for determining the complex of activation regions of a neural network. \cite{huchette2026when} provide a more in-depth overview and survey of this research area.

%% file: arXiv/background.tex
\section{Background} \label{sec:background}

In this section, we recall some general definitions and facts about linear regions of neural networks and their connection to tropical and polyhedral geometry as relevant to our study. 

\subsection{Linear Regions of Neural Networks}
A \emph{neural network} is a function $f:\RR^n\to\RR^m$ of the form
$$
    L_d\circ \sigma_{d-1} \circ L_{d-1} \circ \dotsb\circ\sigma_1\circ L_1
$$
where $L_i:\RR^{n_{i-1}}\to\RR^{n_i}$ is an affine map and $\sigma_i$ is a nonlinear \emph{activation} function. Typically, $\sigma_i$ is computed by applying a univariate function (e.g., $\mathrm{ReLU}$) coordinate-wise.

The number of affine building blocks required to represent a neural network provides a measure of its complexity.
Intuitively, a piecewise linear function that is defined using many affine functions is more complex than one that requires only a few. The idea of ``affine building block'' can be made more precise, and corresponds to the \emph{linear regions} of the function under consideration.

\begin{figure}[ht]
    \centering
    \begin{subfigure}[b]{0.48\textwidth}
        \centering
        \includegraphics[width=0.7\textwidth]{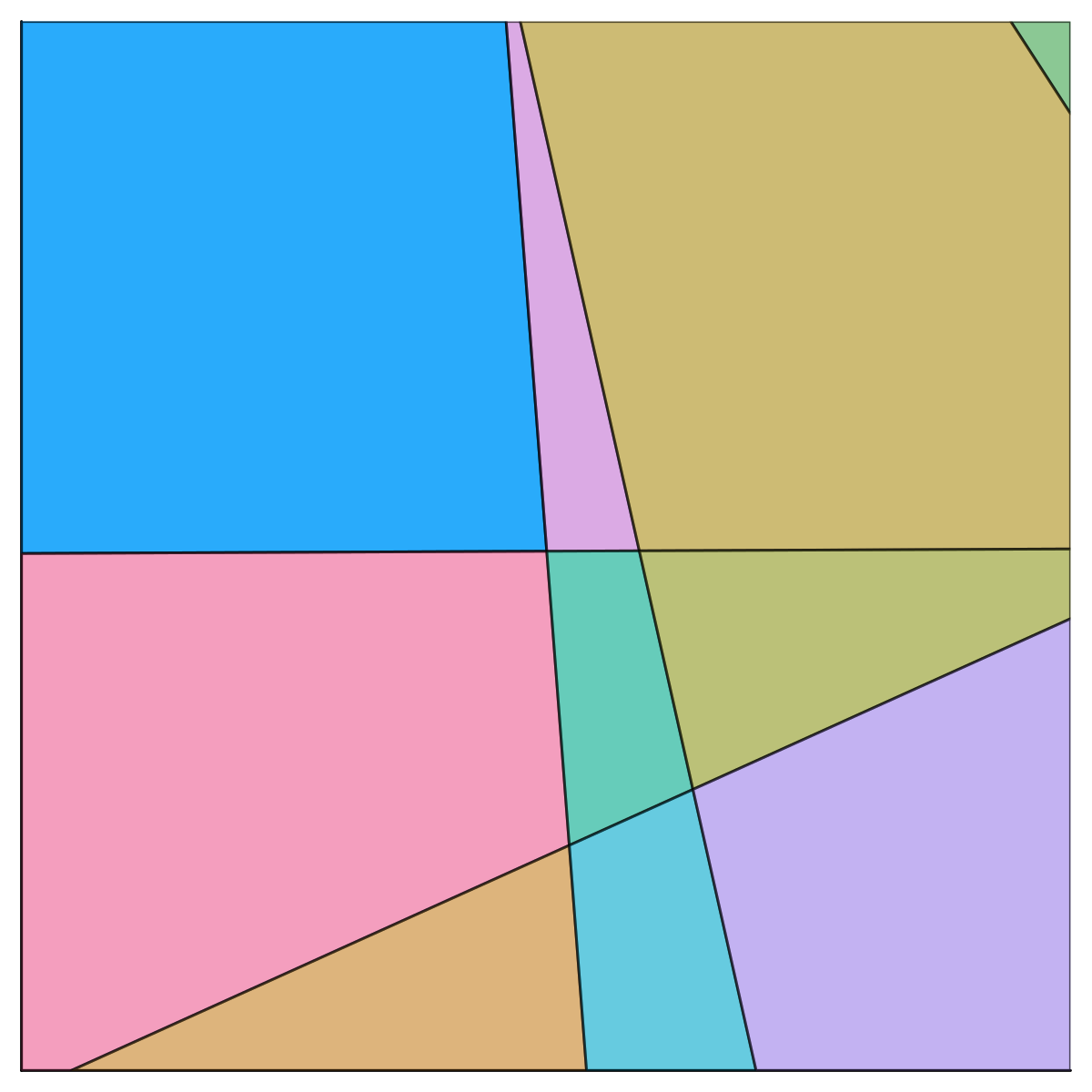}
        \caption*{Narrow Neural Network}
    \end{subfigure}
    \hfill
    \begin{subfigure}[b]{0.48\textwidth}
        \centering
        \includegraphics[width=0.7\textwidth]{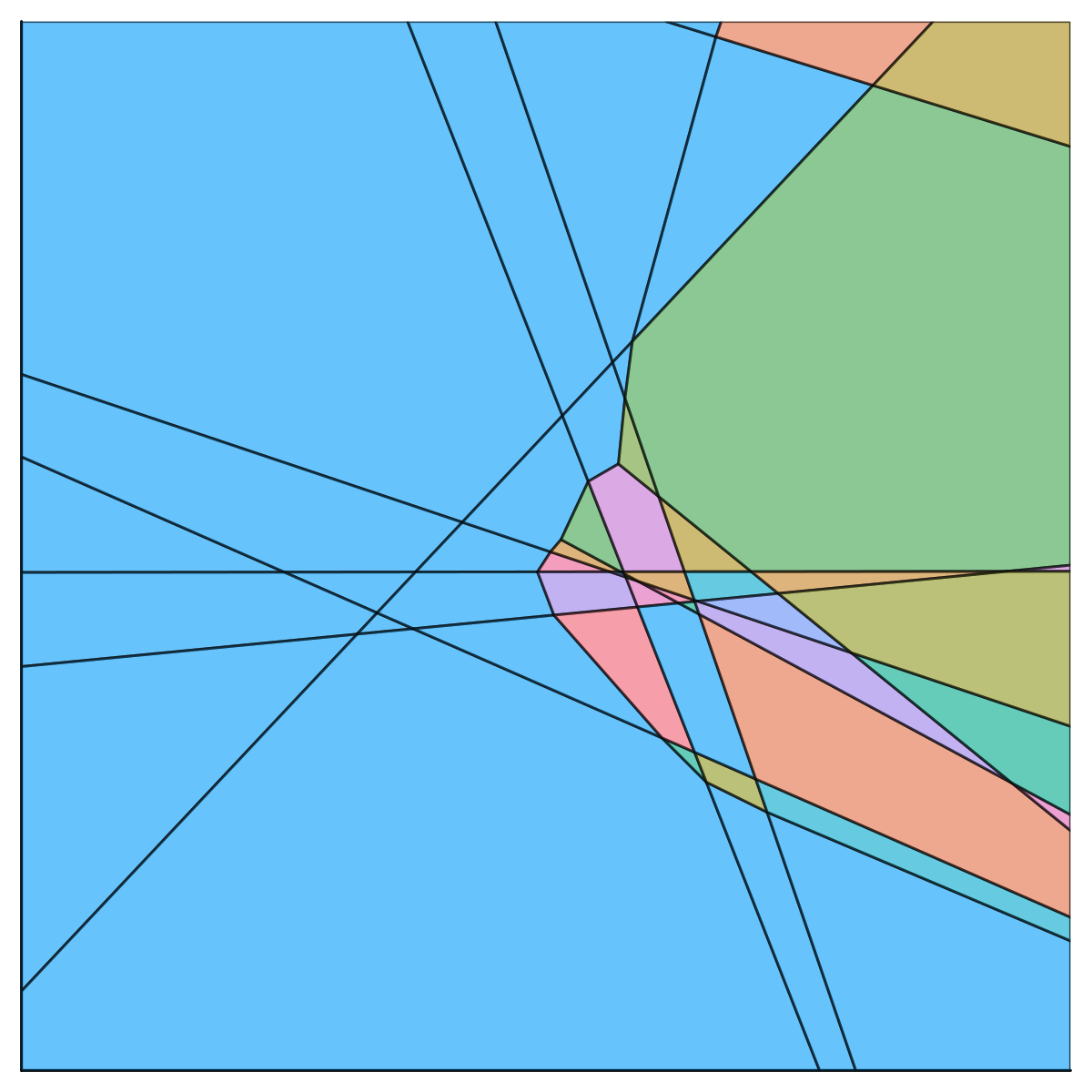}
        \caption*{Wide Neural Network}
    \end{subfigure}
    \caption{
    On the left, we visualize the linear regions of a neural network with width 4.
    On the right, we visualize the linear regions of a relatively larger neural network which has width 12.
    }
    \label{fig:nn_linear_regions}
\end{figure}

\begin{definition} \label[definition]{def:linear-region}
    A set $U\subset\mathbb{R}^n$ is a \emph{linear region} of a neural network $f:\mathbb{R}^n\to\mathbb{R}^m$ if it is a maximal region (closure of a connected non-empty open set) on which $f$ is affine (i.e., it can be represented by an affine map).
\end{definition}

\begin{remark} \label[remark]{rmk:linear-regions-definitions}
    Our definition of a linear region differs from some other works that appear in the literature as it does not allow linear regions to have lower dimensional components. The main motivation behind this choice is that it is consistent with the usual usage of the term ``region'' in mathematics.
    In particular, a connected set on which the network is linear may still account for several linear regions, as illustrated in \cref{fig:single-point-regions}.
    We refer to \cite{stargalla2025computational} for an in-depth comparison of the various definitions that appear in the literature. Note that linear regions according to our definition correspond to closures of open connected linear regions in the sense of \cite{stargalla2025computational}. In particular, we will see that descriptions of open connected linear regions in the same sense can be recovered from the descriptions that our algorithms produce.
\end{remark}

\begin{figure}[ht]
\centering
\definecolor{regionblue}{RGB}{80,179,232}
\definecolor{regionpink}{RGB}{238,165,196}
\definecolor{regiongreen}{RGB}{160,207,169}
\begin{tikzpicture}[scale=0.7]
  \begin{scope}
    \clip (-3,-3) rectangle (3,3);
    \fill[regionblue]  (0.3,-0.4) -- (7.67,4.76) -- (-7.07,4.76) -- cycle;
    \fill[regionblue]  (0.3,-0.4) -- (-7.85,-4.20) -- (4.10,-8.55) -- cycle;
    \fill[regionpink]  (0.3,-0.4) -- (4.10,-8.55) -- (7.67,4.76) -- cycle;
    \fill[regiongreen] (0.3,-0.4) -- (-7.07,4.76) -- (-7.85,-4.20) -- cycle;
    \draw[semithick, black] (0.3,-0.4) -- (7.67,4.76);
    \draw[semithick, black] (0.3,-0.4) -- (-7.07,4.76);
    \draw[semithick, black] (0.3,-0.4) -- (-7.85,-4.20);
    \draw[semithick, black] (0.3,-0.4) -- (4.10,-8.55);
  \end{scope}
  \node[darkblue] at (0.3,1.6) {$U_1$};
  \node[darkblue] at (-0.3,-2.1) {$U_2$};
  \draw[semithick, black] (-3,-3) rectangle (3,3);
\end{tikzpicture}
\caption{The two cones $U_1$ and $U_2$ share the same linear map and meet at their common vertex. Each is therefore a separate linear region according to our definition, but would form a single region from the point of view of other definitions in the literature.}
\label{fig:single-point-regions}
\end{figure}

In the following, we will often refer to nonempty sets that are contained in the closure of their interior as ``full-dimensional.'' In the case of polyhedra $P \subset \RR^n$, this is equivalent to the condition that $\dim P = n$. We will see that the linear regions of neural networks can always be represented as unions of polyhedra, and correspond to connected full-dimensional components of subsets of $\RR^n$ on which the neural network coincides with a given affine map. Note that by definition, if $A$ and $B$ are linear regions and $A$ intersects with the interior of $B$ then $A = B$.

\subsection{Tropical and Polyhedral Geometry}

In the same sense that algebraic geometry studies geometric properties of solution sets of polynomial systems that can be expressed algebraically, such as their degree, dimension, and irreducible components, \emph{tropical geometry} studies polynomials that are defined using \emph{tropical operations}.

\begin{definition}
    The \emph{tropical operations} are defined as follows:
    \begin{align*}
        a \oplus b & = \max(a, b) & & \text{Tropical addition}\\
        a \odot b & = a + b & & \text{Tropical multiplication} \\
        a \oslash b & = a - b & & \text{Tropical division} \\
        a ^ {\odot b} & = a b & & \text{Tropical exponentiation}
    \end{align*}
    The \emph{tropical semiring} is defined as $\mathbb{T} = \mathbb{R} \cup \{-\infty\}$ endowed with tropical addition and multiplication. 
\end{definition}

\begin{definition} \label[definition]{def:tropical-polynomial}
    A \emph{tropical polynomial} in $\vx = (x_1, \dotsc, x_n)$ is a finite sum of the form
    $$
        c_1 \odot \vx^{\odot \va_1} \oplus \dotsb \oplus c_k \odot \vx^{\odot \va_k}
    $$
    where $c_i \in \mathbb{R}, \va_i \in \NN^n$ and
    $$
    \vx^{\odot \va_i} = x_1^{\odot a_{i1}} \odot \dotsb \odot x_n^{\odot a_{in}}.
    $$
\end{definition}

Every tropical polynomial $f = c_1 \odot \vx^{\odot \va_1} \oplus \dotsb \oplus c_k \odot \vx^{\odot \va_k}$ defines a function $\RR^n \to \R$ given by
$$
f(\vx) = \max_i \left(c_i + \langle \va_i, \vx \rangle\right).
$$

We note that the ReLU activation function is an elementary example of a tropical polynomial: in tropical notation, we have $\mathrm{ReLU}(x) = 0 \oplus x$. More generally, as we will see, every neural network with ReLU activation admits a tropical form. The expression in \cref{def:tropical-polynomial} makes sense with non-integer exponents and the following more general notion applies.

\begin{definition} \label{def:sig}
    A \emph{signomial} is a finite sum of the form
    $$
        c_1 \odot \vx^{\odot \va_1} \oplus \dotsb \oplus c_k \odot \vx^{\odot \va_k}
    $$
    where $c_i \in \mathbb{R}, \va_i \in \RR^n.$
\end{definition}
When writing such tropical expressions, we will always implicitly assume that the exponents $\va_1, \dotsc, \va_k$ are distinct. We point out that exponentiation distributes over addition when the exponents are nonnegative. In particular, composing two signomials with nonnegative exponents yields another signomial with nonnegative exponents.

The notion of a rational function also admits a tropical analog, namely \emph{tropical rational functions}, which are simply expressions of the form $p \oslash q$. In the case where $p$ and $q$ are signomials, we refer to such an expression as a \emph{rational signomial}. Throughout this paper, we will assume that the activation functions we are working with can be expressed as \emph{rational signomials}.

An important property of rational signomials is that they are closed under composition: a composition of rational signomials is again a rational signomial. In contrast, ordinary signomials are not closed under arbitrary composition. As before, every tropical rational function or rational signomial
$$
    f(\vx) = \left(c_1 \odot \vx^{\odot \va_1} \oplus \dotsb \oplus c_k \odot \vx^{\odot \va_k} \right) \oslash \left( d_1 \odot \vx^{\odot \vb_1} \oplus \dotsb \oplus d_\ell \odot \vx^{\odot \vb_\ell} \right)
$$
can be viewed as a function
$$
    f(\vx) = \max_i \left(c_i + \langle \va_i, \vx \rangle\right) - \max_j \left(d_j + \langle \vb_j, \vx \rangle\right).
$$
We also note that certain monomials appearing in a signomial (or rational signomial) may be redundant: removing them does not alter the function. For example, in the expression 
$$
f(x) = 0 \oplus x \oplus x^{\odot 2} = \max(0, x, 2x),
$$ 
the monomial $x$ is redundant since we have $f(x) = 0 \oplus x^{\odot 2}.$

We finish by noting that various other activation functions for neural networks admit such tropical representations. For example, the maxout function can be written as
$$
\mathrm{Maxout}(x_1, \dotsc, x_n) = \max\{x_1, \dotsc, x_n\} = x_1 \oplus \dotsb \oplus x_n
$$
and the leaky ReLU function can be written as
$$
\mathrm{LeakyReLU}_\alpha(x) = \max\{\alpha x, x\} = x^{\odot \alpha} \oplus x 
$$
when the constant $\alpha$ satisfies $0 < \alpha < 1$.

\subsection{Tropical Expressions of Neural Networks}

One of the key observations connecting tropical geometry and deep learning is that neural networks can be written as rational signomials. The following result can be seen as a generalization of \citet[Proposition 5.6]{zhang2018tropical}.

\begin{theorem}
\label{thm:nn-rational-signomial-form}
    Every feedforward neural network whose activation functions are vectors of rational signomials can be written as a vector of rational signomials.
\end{theorem}

\begin{proof}
    Let $f:\mathbb{R}^n\to\mathbb{R}^m$ be a neural network.
    Write
    $$
    f = L_d \circ \sigma_{d-1} \circ L_{d-1} \circ \dotsb \circ \sigma_1 \circ L_1,
    $$
    and 
    $$
    L_k(\vx) = A^{(k)} \vx + b^{(k)}.
    $$
    Note that
    $$
    (L_k(\vx))_j = b^{(k)}_j \odot x_1^{\odot A^{(k)}_{j1}} \odot \dotsb \odot x_{n_{k-1}} ^ {\odot A^{(k)}_{jn_{k-1}}}  
    $$
    Thus, every component of $L_k(\vx)$ can be written as a rational signomial in $\vx$.
    In particular, $f$ can be written as a composition of vectors of rational signomials, so $f$ must be a vector of rational signomials.
\end{proof}

The proof that neural networks can be written as rational signomials provides a recursive construction that we will use to computationally obtain a tropical representation of a neural network. The proof also shows that any such network can be written as a composition of rational signomials, which will allow us to compute their linear regions further on, using \cref{alg:layerwise-linear-regions}.

\begin{remark} \label[remark]{rmk:nonneg-exponents}
    If the neural network has nonnegative weights and all activation functions are signomials with nonnegative exponents, then the proof above shows that the network can be represented by a vector of signomials (rather than rational signomials) with nonnegative exponents.
\end{remark}

\subsection{Polyhedra and Linear Regions} \label{subsec:linear-regions-signomials}
The study of tropical objects is intimately connected to polyhedral geometry, as their linear regions can be expressed as polyhedra or unions of polyhedra. In this section, we recall some basic facts about these objects. 

\begin{definition}
    A \emph{polyhedron} is a subset of $\mathbb{R}^n$ of the form $P(\mA,\vb) := \{\vx\in\mathbb{R}^n:\mA\vx\leq\vb\},$
    where $\mA\in\mathbb{R}^{m\times n}$, $\vb\in\mathbb{R}^m$, and the inequality is taken element-wise.
\end{definition}

\begin{definition}
    The \emph{dimension} of a polyhedron $P$ is the dimension of the smallest affine subspace of $\RR^n$ that contains $P$. This subspace is called the \emph{affine hull} of $P$, and denoted $\mathrm{AffHull}(P).$ If $P$ is empty, we follow the convention that $\dim P = -1$.
\end{definition}

\begin{example} The following are polyhedra:
\begin{itemize}
    \item Any line $L = \{ t \cdot \vu + \vv \mid t \in \RR\}$ is a polyhedron. The affine hull of $L$ is simply $L$. In particular, the dimension of $L$ as a polyhedron is equal to 1.
    \item Any vector subspace of $\RR^n$, and more generally any affine subspace of $\RR^n$, is a polyhedron. The affine hull of an affine subspace $A \subset \RR^n$ is simply $A$. Thus the dimension of $A$ as a polyhedron coincides with the dimension of $A$ as an affine subspace.
    \item The triangle $T$ described by
    $$
        T = \{ (x, y) \mid x \ge 0, y \ge 0, x + y \le 1\}
    $$
    is a polyhedron. The affine hull of $T$ is $\RR^2$, so $\dim T = 2$.
\end{itemize}
\end{example}

We now show how the linear regions of a signomial can be expressed as polyhedra. Let $f$ be a signomial in $n$ variables. We write
\begin{align*}
    f(\vx) & = c_1 \odot \vx^{\odot \va_1} \oplus \dotsb \oplus c_k \odot \vx^{\odot \va_k} \\
       & = \max_i \left(c_i +\langle\va_i, \vx \rangle\right).
\end{align*}
For $i\in\{1,\dots,k\}$, we can consider the set of points $\vx$ where $f$ is equal to its $i$th monomial, that is:
$$
U_i=\left\{\vx\in\mathbb{R}^n:f(\vx)=c_i+\langle\va_i,\vx\rangle\right\},
$$
A brief calculation shows that
$
U_i = P(\mM_i, \vh_i)
$
where $\mM_i \in \RR^{k \times n}$ and $\vh_i \in \RR^k$ are given by
$$
    \mM_i = \begin{pmatrix}
        a_{11} - a_{i1} & \dotsb & a_{1n} - a_{in} \\
        \vdots & \ddots & \vdots \\
        a_{k1} - a_{i1} & \dotsb & a_{kn} - a_{in}
    \end{pmatrix} \\
    = \mA - \mathbf{1}\va_i^{\top}
$$
and
$$
\vh_i = \begin{pmatrix}
    c_i - c_1 \\
    \vdots \\
    c_i - c_k
\end{pmatrix} =
c_i \mathbf{1} - \mathbf{c}.
$$
Here, $\mA$ is the $k\times n$ matrix whose rows are $\va_1^{\top},\dotsc,\va_k^{\top}$, $\vc = (c_1, \dotsb, c_k)^{\top}$ and $\mathbf{1} =(1, \dotsc, 1)^{\top}$.
It is then possible to check that the set of linear regions is a subset of $\{U_1, \dotsc, U_k\}$, namely the set of all $U_i$ with dimension $n$. In particular, some of the $U_i$ may not actually be linear regions: these correspond to the redundant monomials in the expression of $f$.

\begin{example}\label{eg:signomials}
    \begin{enumerate}
        \item[]
        \item Consider the univariate signomial $p(x)=0\oplus 1 \odot x \oplus 1 \odot x^2$. Then the map $p$ is given by
        $$
        x\mapsto\max\left\{0,1+x,1+2x\right\},
        $$
        thus
        $$
        \begin{cases}
            U_0=P(1,-1)=\{x\le-1\},\\
            U_1=P\left(\begin{bmatrix} 1 \\ -1\end{bmatrix},\begin{bmatrix}0\\1\end{bmatrix}\right)=\{-1\leq x\leq0\},\\U_2=P(-1,0)=\{x\ge0\}.
        \end{cases}
        $$
        The linear regions of $p$ are then $\{U_0,U_1,U_2\}$.
        \item Consider the signomial $p=1\odot x\oplus0 \odot x^2$. Then the map $p$ is given by 
        $$
        x\mapsto\max\{1+x,2x\},
        $$
        thus
        $$
        \begin{cases}
            U_1=P(1,1)=\{x\leq1\},\\
            U_2=P(-1,-1)=\{x\geq1\}.
        \end{cases}
        $$
        The linear regions of $p$ are then $\{U_1,U_2\}$.
        \item Consider the signomial $p=1 \odot x^0\oplus 1 \odot x \oplus 1 \odot x ^2$. Then the map $p$ is given by 
        $$
        x\mapsto\max\{1,1+x,1+2x\},
        $$
        thus
        $$
        \begin{cases}U_0=P(1,0)=\{x\leq0\},\\
            U_1=P\left(\begin{bmatrix}1 \\ -1\end{bmatrix},\begin{bmatrix}0\\0\end{bmatrix}\right)=\{x=0\},\\
            U_2=P(-1,0)=\{x\geq0\}.
        \end{cases}
        $$
        The linear regions in this case are $\{U_0,U_2\}$. Note how $U_1$ does not contribute to the map $p$, so the monomial $1\odot x$ is redundant. Redundancy in this case can be detected by noting that the dimension of $U_1$ is less than the number of variables of $p$ (see \cref{lem:redundant-monomial} and \cref{alg:pruned-expression}).
        \item Consider the signomial $p=1\odot x^2\oplus 1\odot x^3\oplus 2\odot x^4$. Then the map $p$ is given by
        $$
        x\mapsto\max\{1+2x,\, 1+3x,\, 2+4x\},
        $$
        thus
        $$
        \begin{cases}U_2=P\left(1,-\frac{1}{2}\right)=\left\{x\leq-\frac{1}{2}\right\},\\U_3=\varnothing\\U_4=P\left(-1,\frac{1}{2}\right)=\left\{x\geq-\frac{1}{2}\right\}.\end{cases}
        $$
        Hence, the linear regions in this case are $\{U_2,U_4\}$. Note how $U_3$ does not contribute to the map $p$. Hence, the monomial $1\odot x^3$ is redundant.
    \end{enumerate}
\end{example}

For a signomial $f$, a point that lies in two linear regions is a point at which the maximum defining $f$ is attained at two different terms in the expression of $f$. In standard tropical geometry terminology, these points are precisely those found on the tropical hypersurface cut out by $f$. 

We finish by noting that this discussion extends to higher dimensions: if $\vp = (p_1, \dotsc, p_m)$ is a vector of signomials, and the linear regions of $p_i$ are given by $U_{i1}, \dotsc, U_{ij_i}$ then the linear regions of $\vp$ are given by intersections of the form $U_{1k_1} \cap \dotsb \cap U_{mk_m}$. This can be summarized by \cref{alg:linear-regions-signomial}, and we give a proof of the correctness of this algorithm in \cref{prop:linear-regions-signomial}. Before doing so, we will need the following lemma about refinements of polyhedral subdivisions.

\begin{lemma}\label[lemma]{lem:polyhedral-covering}
    Suppose we are given finite polyhedral subdivisions $\{P_{1j}\}, \dotsc, \{P_{mj}\}$, i.e., finite collections of polyhedra such that for all $i$, (i) the $\{P_{ij}\}$ cover $\RR^n$ and (ii) the $\{P_{ij}\}$ do not overlap (i.e., $P_{ij} \cap P_{ik} \subset \partial P_{ij} \cap \partial P_{ik}$). Then the collection of full-dimensional intersections of the form
    $$
        P_{1k_1} \cap \dotsb \cap P_{mk_m}
    $$
    is also a polyhedral subdivision. 
\end{lemma}

\begin{proof}
    We first show that if $\{Q_i\}$ is a finite collection of polyhedra that satisfies the properties (i) and (ii) above then the subcollection
    $$
    \{ Q_i \mid \dim Q_i = n\}
    $$
    also satisfies these properties.
    Property (ii) is satisfied, so we only need to check property (i).
    Let $x$ be a point that does not lie in a full-dimensional $Q_i$. Then, since the $Q_i$ are closed, there is an open $U$ containing $x$ that does not intersect with the full-dimensional $Q_i$. This implies that $U$ is contained in the union of finitely many polyhedra of codimension at least 1, which is a contradiction.
    The result then follows from applying this to the collection of intersections
    $$
    P_{1k_1} \cap \dotsb \cap P_{mk_m}
    $$
\end{proof}

\begin{algorithm}
    \begin{algorithmic}[1]
        \Require $\vp = (p_1, \dotsc, p_m)$ a vector of signomials.
        \State Compute the set $\{ U_{i1}, \dotsc, U_{ij_i}\}$ of linear regions of $p_i$
        \State Set $L = \emptyset$
        \For{tuples of indices $(i_1, \dotsc, i_m)$}
            \State Compute the intersection $U_{i_1\dotsb i_m} = U_{1i_1} \cap \dotsc \cap U_{mi_m}$
            \If{$U_{i_1\dotsb i_m}$ is full-dimensional}
            \State Add $U_{i_1\dotsb i_m}$ to $L$
            \EndIf
        \EndFor
        \Return $L$
    \end{algorithmic}
    \caption{Computation of linear regions for vectors of signomials}
    \label{alg:linear-regions-signomial}
\end{algorithm}

\begin{proposition}\label[proposition]{prop:linear-regions-signomial}
    \cref{alg:linear-regions-signomial} computes the linear regions of a vector of signomials.
\end{proposition}

\begin{proof}
    Let $\vp=(p_1, \dotsc, p_m)$ be a vector of signomials, and let $\{U_{i1}, \dotsc, U_{ij_i}\}$ be the set of linear regions of signomial $p_i$. We need to show that the full-dimensional intersections of the form
    $$
    U_{1k_1} \cap \dotsb \cap U_{mk_m}
    $$
    are the linear regions of $\vp$. By \cref{lem:polyhedral-covering}, these intersections cover $\RR^n$. We have that $\vp$ is affine on each of these sets, so it suffices to check maximality. Suppose we can find a connected open set $T$ such that the closure $\overline T$ contains $U_{1k_1} \cap \dotsb \cap U_{mk_m}$ strictly, and such that $\vp$ is affine on $\overline T$. Then it must be the case that $T$ intersects with the interior of some intersection $U_{1\ell_1} \cap \dotsb \cap U_{m\ell_m}$, where for some index $i$, we have $\ell_i \ne k_i$. 
    In particular, for that index $i$, $p_i$ is represented by the same affine map on $U_{ik_i}$ and on $U_{i\ell_i}$, which is a contradiction (for signomials, each linear region corresponds to a different affine map).
\end{proof}

%% file: arXiv/expressivity.tex
\section{Tropical Measures and Invariants for Neural Networks} \label{sec:expressivity}

In this section, we present our main theoretical results on the linear regions of rational signomials and neural networks. We start by presenting an algorithm that determines the exact linear regions of neural networks. We then introduce the \emph{Hoffman constant of a rational signomial}, and show how it relates to the distance of linear regions to the origin.
Finally, we study the monomials that appear in tropical expressions of neural networks and show how they can be viewed as an algebraic measure of the complexity of such tropical expressions.

\subsection{Linear Regions of Rational Signomials} \label{subsection:rational-signomial-linear-regions}

\input{arXiv/subsec_lr_new}

\subsection{Hoffman Constants for Neural Network Linear Regions}

Many methods for studying the linear regions of a neural network operate on a bounded subset of the input space. This raises a natural geometric question: how large must such a domain be in order to guarantee that every linear region of the network intersects with it?

We address this question using Hoffman constants, which are classical error-bound constants for systems of linear inequalities \citep{hoffman1952approximate}. Our contribution is to connect them to the polyhedral subdivisions induced by tropical representations of neural networks and use them to obtain quantitative control over the distance from the network's linear regions to the origin. In particular, we derive a computable upper bound on the radius of an $\ell^\infty$-ball guaranteed to intersect every linear region.

\paragraph{Motivation: The Minimal Effective Radius.}

Various approaches for computing the number of linear regions of a neural network $f$ restrict to counting the number of regions that intersect with some bounded domain $[-R, R]^n$. In order to choose an appropriate $R$, the maximum distance of a linear region to the origin is required.
Beyond merely counting the number of linear regions of a rational signomial, there are other quantities of interest that arise from viewing neural networks as piecewise linear functions. For example, it would be interesting to study how the distance from linear regions to data points used during training varies.

Given a neural network $f$ with linear regions $\mathcal{U} = \{U_1,\ldots,U_m\}$, we define the \emph{minimal effective radius} of $f$ at $\vx \in \RR^n$ as
$$
R_f(\vx):=\min\left\{r: \text{for all } U\in\mathcal{U}
, \bar{B}(\vx,r)\cap U \neq \varnothing\right\}
$$
where $\bar{B}(\vx,r)$ is the closed ball of radius $r$ centered at $\vx$ with respect to the $\| \cdot \|_\infty$ norm. That is, $R_f(\vx)$ is the minimal radius such that the ball $\bar{B}(\vx,r)$ intersects all linear regions. This measures the maximal distance from a point $\vx$ to a linear region of $f$. In particular, when $\vx=\mathbf 0$,
the box $[-R_f(\mathbf 0),R_f(\mathbf 0)]^n$ intersects every linear region of the network. We use Hoffman constants associated with the polyhedral systems induced
by a tropical representation of a neural network to study the quantity
$R_f(\vx)$.

\paragraph{The Hoffman Constant of a Matrix.} Intuitively, the Hoffman constant of a matrix measures the distance from points in a space to polyhedra constructed from this matrix.
Let $\mA$ be an $m\times n$ matrix with real entries and suppose we have fixed norms on both $\RR^n$ and $\RR^m$. Then for any $\vb\in\mathbb{R}^m$ such that $P(\mA, \vb)$ is non-empty, let
$$
d(\vu,P(\mA, \vb))=\min\{\|\vu- \vx\|:\vx\in P(\mA, \vb)\}
$$
denote the distance of a point $\vu\in\mathbb{R}^n$ to the polyhedron. There exists a minimal constant $H_{\|\cdot\|}(A)$ \emph{that depends only on $A$} such that
\begin{equation}\label{eq:hoff-fundamental}
    d(\vu,P(\mA, \vb))\le H(\mA)\left\|(\mA \mathbf{u}-\vb)_+\right\|
\end{equation}
for all  $\vu\in\mathbb{R}^n$, where $\vx_+=\max(\vx,0)$ is applied coordinate-wise \citep{hoffman1952approximate}. The number $H(\mA)$ is the \emph{Hoffman constant} of $\mA$. Note that this depends on the choice of norms on $\RR^n$ and $\RR^m$. In our case, we will always work with the $\ell^\infty$ norm on both spaces.

\paragraph{A Hoffman Constant for Rational Signomials.} As we saw in the previous section, rational signomials naturally induce a decomposition of their input space into non-overlapping polyhedra, which glue together to form the linear regions of the signomial. We can use this to extend the notion of a Hoffman constant to signomials.

Let $$p = c_1 \odot \vx^{\odot \va_1} \oplus \dotsb \oplus c_k \odot \vx^{\odot \va_k} \quad \mbox{and} \quad q = d_1 \odot \vx^{\odot \vb_1} \oplus \dotsb \oplus d_\ell \odot \vx^{\odot \vb_\ell}$$ be tropical signomials.
As before, we write $\mA$ for the matrix whose rows are $\va_1^{\top},\dotsc,\va_k^{\top}$, $\vc = (c_1, \dotsb, c_k)^{\top}$, and similarly for $\mathbf{B}$ and $\mathbf{d}$.

We recall from \cref{subsec:linear-regions-signomials} that the linear regions of $p$ correspond to the polyhedra of the form $U_i = P(\mM_i, \vh_i)$
that are full dimensional, where
$$
\mM_i = \mA - \mathbf 1\va_i^{\top}.
$$
Similarly the linear regions of $q$ correspond to polyhedra of the form
$V_j = P(\mN_j, \vk_j)$ that are full dimensional, where
$$
\mN_j = \mB - \mathbf 1\vb_j^{\top}.
$$
As we saw previously, the linear regions of $p \oslash q$ are unions of polyhedra of the form
$$
U_i \cap V_j = P\left(\begin{pmatrix}
            \mM_i \\
            \mN_j
        \end{pmatrix}, \begin{pmatrix}
            \vh_i \\
            \vk_j
        \end{pmatrix} \right),
$$
and we have
$$
d(\vx, U_i \cap V_j) \le H\left(\begin{pmatrix}
            \mM_i \\
            \mN_j
        \end{pmatrix} \right) \left\Vert \left(
        \begin{pmatrix}
            \mM_i \\
            \mN_j
        \end{pmatrix} \vx - \begin{pmatrix}
            \vh_i \\
            \vk_j
        \end{pmatrix} \right)_+ \right\Vert.
$$
In order to bound the distance from a point $\vx$ to the farthest linear region from it, we need to take the maximum over all Hoffman constants corresponding to such polyhedra. This yields the following definition.

\begin{definition}\label[definition]{def:hoff_constant_trop_rat_map}
     The \emph{Hoffman constant of the pair of signomials} $(p, q)$ is
    \begin{equation}\label{eq:def-hoff-rational-map}
        H(p, q) = \max_{\substack{i, j \text{ s.t} \\ \dim (U_i \cap V_j) = n}} H\left(\begin{pmatrix}
            \mM_i \\
            \mN_j
        \end{pmatrix} \right).
    \end{equation}

\end{definition}

Since every neural network $f$ in our setting admits a rational signomial representation, \cref{def:hoff_constant_trop_rat_map} associates a Hoffman constant with any chosen tropical representation $ f = p \oslash q$. It is important to emphasize that this quantity is associated with the representation $(p,q)$ rather than being an intrinsic invariant of the underlying function.

Given a signomial $p(\vx)=\max_i \left(\langle \va_i, \vx \rangle + c_i\right)$, we define the \emph{min-conjugate of $p$} as $\check{p}(\vx)=\min_i \left(\langle \va_i, \vx \rangle + c_i\right)$. We can then relate the minimal effective radius and the Hoffman constant of a rational signomial; the following result provides the key connection between this representation-level Hoffman constant and the geometry of the neural network's linear regions.

\begin{proposition}
\label[proposition]{prop:effective-radius}
    Let $f = p\oslash q$ be a rational signomial. For any $\vx\in\mathbb{R}^n$, we have
    \begin{equation}\label{eq:rf-hpq}
       R_f(\vx)\le H(p,q)\max\{p(\vx)-\check{p}(\vx),\, q(\vx)-\check{q}(\vx)\}.
    \end{equation}
\end{proposition}

\begin{proof}
The polyhedra defined by 
$$
\begin{pmatrix}
            \mM_i \\
            \mN_j
        \end{pmatrix} \vx \le \begin{pmatrix}
            \vh_i \\
            \vk_j
        \end{pmatrix}
$$
of dimension $n$ form a convex refinement of linear regions of $f$. Let 
$$\mathrm{res}_{i,j}(\vx):=\begin{pmatrix}
            \mM_i \\
            \mN_j
        \end{pmatrix} \vx- \begin{pmatrix}
            \vh_i \\
            \vk_j
        \end{pmatrix}
$$        
denote the residual of $\vx$ to the polyhedron. We have 
$$
R_f(\vx)\le H(p,q)\max\left\{\|\mathrm{res}_{i,j}(\vx)_+\|_\infty: 1 \le i \le k, 1 \le j \le \ell, \dim U_i \cap V_j = n \right\}.
$$
Note that
\begin{align*}
    \|\mathrm{res}_{i,j}(\vx)_+\|_\infty
        &= \left\|\left(
            \begin{bmatrix}\mA \vx+\vc-\mathbf{1}(\langle \va_i, \vx \rangle+c_i)\\
                            \mB \vx+\vd-\mathbf{1}(\langle \vb_j, \vx \rangle+d_j)\end{bmatrix}\right)_+\right\|_\infty\\
        &=\max_{r,s}\left\{(\mA \vx+\vc)_r-(\langle \va_i, \vx \rangle+c_i),\,(\mB \vx+\vd)_s-(\langle \vb_j, \vx \rangle+d_j),\, 0\right\}\\
        &=\max\left\{p(\vx)-(\langle \va_i, \vx\rangle+c_i),\, q(\vx)-(\langle \vb_j, \vx\rangle + d_j),\, 0\right\}
\end{align*}
Therefore,
\begin{align*}
    \max_{\substack{i, j \text{ s.t} \\ \dim (U_i \cap V_j) = n}}\|\mathrm{res}_{i,j}(\vx)_+\|_\infty
        &=\max_{\substack{i, j \text{ s.t} \\ \dim (U_i \cap V_j) = n}}\left\{p(\vx)-(\langle \va_i, \vx\rangle +c_i),\, q(\vx)-(\langle \vb_j,  \vx \rangle+d_j),\, 0\right\}\\
        &\le\max\left\{p(\vx)-\min_i\{\langle \va_i, \vx \rangle +c_i\},\, q(\vx)-\min_j\{\langle \vb_j, \vx \rangle +d_j\},\, 0\right\}\\
        &=\max\left\{p(\vx)-\check{p}(\vx),\,q(\vx)-\check{q}(\vx)\right\}
\end{align*}
which proves \eqref{eq:rf-hpq}.
\end{proof}

\begin{corollary}
Let $f = p \oslash q$ be a rational signomial representation of a neural network and let 
$$
R > H(p,q) \max \{ p(\mathbf 0) - \check p(\mathbf 0),\, q(\mathbf 0) - \check q(\mathbf 0) \}.
$$
Then every linear region of $f$ intersects the box $[-R,R]^n$.
\end{corollary}

\begin{proof}
By \cref{prop:effective-radius},
$$
R_f(\mathbf 0) \le H(p,q) \max\{p(\mathbf 0)- \check p(\mathbf 0),\, q(\mathbf 0) - \check q(\mathbf 0)\} < R.
$$
Hence the closed $\ell^\infty$-ball of radius $R$ centered at the origin, which is precisely $[-R,R]^n$, intersects every linear region of $f$.
\end{proof}

More generally, given vectors of signomials $\vp = (p_1, \dotsc, p_m), \vq =  (q_1, \dotsc, q_m)$ in $n$ variables, we can generalize \cref{def:hoff_constant_trop_rat_map} to the vector pair $(\vp,\vq)$. Indeed, let $\{U_i = P(\mathbf M_i, \mathbf h_i),i\in I\}$ be the linear regions obtained by \cref{alg:linear-regions-signomial} for $\vp$, and similarly $\{V_j = P(\mathbf N_j, \mathbf k_j),j\in J\}$ are the linear regions for $\vq$. Then the Hoffman constant $H(\vp,\vq)$ is defined in a similar way as \Cref{def:hoff_constant_trop_rat_map} by maximizing over pairs $(i, j) \in I \times J$ such that $U_i \cap V_j$ is full-dimensional.

\paragraph{Computing Hoffman Constants.}\label{sec:theory-hoffman-computation} 

To make the preceding bounds effective, we require the classical Hoffman constants that appear in \cref{eq:def-hoff-rational-map}. We use the algorithm of \citet{pena2018algorithm}, which we refer to as the \emph{Pe\~{n}a--Vera--Zuluaga (PVZ) algorithm}.

For $\mA\in\mathbb{R}^{m\times n}$ and $J\subseteq \{1,2,\ldots,m\}$, let $\mA_J$ denote the submatrix of $\mA$ consisting of rows with indices in $J$. Following \citet{pena2018algorithm}, the set $J$ is called \emph{$\mA$-surjective} if
$$
    A_J \RR^n + \RR^J_+ = \RR^J.
$$
Let $\mathcal{S}(\mA)$ denote the collection of all $\mA$-surjective sets. The PVZ algorithm is based on the following characterization of the Hoffman constant.

\begin{proposition}{\cite{pena2018algorithm}}\label[proposition]{prop:pvz}
Let $\mA\in\mathbb{R}^{m\times n}$. Equip both $\mathbb{R}^m$ and $\mathbb{R}^n$ with the $\ell^\infty$ norm. Then
\begin{equation}\label{eq:pvz-total}
    H(\mA) = \max_{J\in\mathcal{S}(\mA)} H_J(\mA)
\end{equation}
where $H_{\varnothing}(\mA)=0$ and, for every nonempty $J\in\mathcal{S}(\mA)$,
\begin{equation}\label{eq:pvz-J}
H_J(\mA) = \max_{\substack{\vy\in\mathbb{R}^m \\ \|\vy\|_\infty\le 1}}\min_{\substack{\vx\in\mathbb{R}^n\\ \mA_J\vx\le \vy_J}}\|\vx\|_\infty = \frac{1} {\displaystyle\min_{\substack{\vv\in\mathbb{R}^J_+, \|\vv\|_1=1}}\left\|\mA_J^\top \vv\right\|_1}.
\end{equation}
\end{proposition}

The dual of the $\ell^\infty$ norm is the $\ell^1$ norm. Hence, testing $\mA$-surjectivity and evaluating \cref{eq:pvz-J} can be formulated as linear programming problems (LP). In particular, a nonempty set $J$ is $\mA$-surjective precisely when the optimal value of
\begin{equation}\label{eq:LP}
  \min\left\|\mA^\top_Jv\right\|_1,\,\, s.t.\,\, \vv\in\mathbb{R}^J_+,\|\vv\|_1=1
\end{equation}
is positive. Exact evaluation of \cref{eq:pvz-total} requires a combinatorial search over candidate subsets together with repeatedly solving such LPs. Although the individual LPs are typically inexpensive, their number can make the exact computation costly in practice. We therefore consider lower and upper bounds in order to reduce different aspects of this computation.

\paragraph{Lower and Upper Bounds.}

A lower bound follows directly from \cref{eq:pvz-total}. Rather than searching over the complete collection $\mathcal{S}(\mA)$, we sample candidate subsets $J\subseteq\{1,\ldots,m\}$, test them for $\mA$-surjectivity using \cref{eq:LP} and retain the largest value of
$H_J(\mA)$ encountered. Since the maximum is taken over only a subset of the $\mA$-surjective sets, the resulting quantity is necessarily a
lower bound on $H(\mA)$.

\begin{algorithm}[ht]
  \caption{Randomized lower bound for the Hoffman constant}
    \begin{algorithmic}[1]
        \Require $\mA$: an $m\times n$ matrix, $B$: number of iterations
        \State Initialize $H_L=0$.
        \For{$i\in\{1,\dots,B\}$}
            \State Sample a random integer $K \in \{1, \dotsc, m\}$.
            \State Sample a random subset $J$ of $\{1,\dots,m\}$ of size $K$.
            \State Solve \eqref{eq:LP}. Let $t$ be the optimal value;
            \If{$t>0$}
            \State Update $H_L \gets \max\left\{H_L,\frac{1}{t}\right\}$;
            \Comment{$J$ is surjective.}
            \EndIf
        \EndFor
        \State \Return $H_L$ \Comment{$H_L$ is a lower bound for Hoffman constant.}
    \end{algorithmic}
    \label{alg:lower_hoffman}
\end{algorithm}

For an upper bound, we use an alternative characterization of the Hoffman constant that replaces $\mA$-surjective subsets by full row-rank submatrices. Let $\mathcal{D}(\mA)$ be the set of subsets $J\subseteq \{1,\ldots,m\}$ such that $\mA_J$ has full row rank. Then a standard characterization of the Hoffman constant
\citep{pena2018algorithm} gives
$$
H(\mA) = \max_{\substack{J \in \mathcal D(\mA)}}\frac{1}{\min_{\substack{\vv\in\mathbb{R}^J_+, \|\vv\|_1=1}}\left\|\mA_J^\top \vv\right\|_1}.
$$

Let $J \in \mathcal{D}(\mA)$ and let $\hat \rho(\mA_J)>0$ denote the smallest singular value of $\mA_J$ (equivalently, its smallest nonzero singular value). For any
$\vv \in \RR_+^J$ with $\|\vv\|_1=1$, we have
$$
\|\mA_J^\top\vv\|_1 \geq \|\mA_J^\top\vv\|_2 \geq
\hat\rho(\mA_J)\|\vv\|_2 \geq \frac{\hat\rho(\mA_J)}{\sqrt{|J|}}.
$$
Therefore,
\begin{equation}
\label{eq:hoff-upper}
    H(\mA) \leq \max_{\substack{J\in\mathcal{D}(\mA)\\J\neq\varnothing}}
    \frac{\sqrt{|J|}}{\hat\rho(\mA_J)}.
\end{equation}
This bound is analogous to the one obtained by \cite[Theorem 4.2]{guler1995approximations} with respect to the $\ell^2$-norm.

Unlike the randomized lower bound, computing the right-hand side of \cref{eq:hoff-upper} still requires iterating over a possibly large set $\mathcal{D}(\mA)$. Instead, it replaces solving a linear program by a less expensive singular-value computation. The final procedure is summarized in \cref{alg:upper_hoffman}.

\begin{algorithm}[ht]
  \caption{Upper bound for Hoffman constant}
    \begin{algorithmic}[1]
        \Require $\mA$: an $m\times n$ matrix
        \State Initialize $H_U=0$.
        \For{$J\in\mathcal{D}(\mA),\ J\ne\varnothing$}
            \State Compute $\hat\rho(\mA_J)$, the minimal singular value of $\mA_J$.
            \State Update $H_U=\max\left\{H_U,\frac{\sqrt{|J|}}{\hat\rho(\mA_J)}\right\}$;
        \EndFor
        \State \Return Upper bound for Hoffman constant $H_U$.
    \end{algorithmic}
    \label{alg:upper_hoffman}
\end{algorithm}

\subsection{Monomials} \label{subsec:monomials}

The tropical representation of a neural network provides an algebraic perspective on its complexity that complements the geometric information encoded by its linear regions. In particular, a rational signomial
representation is built from finitely many monomials, and their number provides a natural measure of the complexity of the corresponding symbolic expression. This motivates us to study \emph{monomial complexity}: the number of non-redundant monomials appearing in a tropical representation of a neural network. Unlike the number of linear regions, however, monomial
complexity is a property of the chosen representation rather than an invariant of the underlying piecewise linear function.

\begin{definition}
    Let $f$ be a scalar-valued neural network, and $g \oslash h$ a tropical representation of $f$, i.e., a rational signomial that is equal to $f$. If $g$ has $s$ non-redundant monomials and $h$ has $t$ non-redundant monomials then we define the \emph{monomial complexity} of the representation $g \oslash h$ to be the pair $(s, t)$.
\end{definition}

Intuitively, this quantity captures how many linear terms are needed to express the neural network.
Notice that this measure is closely related to, but not identical to, the number of linear regions of a neural network.

\begin{example}
    \begin{enumerate}
        \item[]
        \item If $g$ is a signomial then the number of non-redundant monomials of $g$ is equal to the number of linear regions of $g$. See \Cref{prop:monomials-linear-regions}.
        \item If $f$ is a neural network with a tropical representation of monomial complexity $(s, t)$ then $f$ has at most $st$ linear regions.
    \end{enumerate}
\end{example}

The number of linear regions and the number of monomials in a tropical expression are closely related, but they capture different notions of complexity. Linear regions are a geometric property of the underlying piecewise linear function, whereas monomial counts describe the algebraic complexity of a particular tropical representation. In \cref{sec:experiments}, we investigate this relationship computationally
for a range of tropical expressions.

To use monomial counts in practice, we must first identify monomials that do not contribute to the represented function. For signomials, this redundancy admits a simple geometric characterization in terms of the polyhedra introduced in \cref{subsec:linear-regions-signomials}.

Let
$$
p(\vx) = c_1 \odot \vx^{\odot \va_1} \oplus \dotsb \oplus c_k \odot \vx^{\odot \va_k}
$$
be a signomial in $n$ variables. Recall that each monomial
$c_i\odot\vx^{\odot\va_i}$ determines a polyhedron $U_i\subseteq\RR^n$
such that, in the expression
\begin{equation}\label{eq:polytope_expression}
    p(\vx) = \max_i \left( c_i+\left\langle\va_i,\vx\right\rangle
    \right),
\end{equation}
the maximum is attained by the $i$th term precisely on $U_i$. The following result characterizes exactly when such a monomial is
redundant.

\begin{proposition} \label[proposition]{lem:redundant-monomial}
    Let $p$ be a signomial as above. Then the $i$th monomial can be removed from the expression of $p$ without changing the underlying function if and only if $\dim U_i < n$.
\end{proposition}

\begin{proof}
Recall that $U_k$ is defined by the system of inequalities
\begin{equation}
\label{eq:monomial-region}
    \langle \va_j, \vx\rangle + c_j \leq \langle \va_k, \vx \rangle + c_k, \qquad j\neq k.
\end{equation}

Suppose first that $\dim U_k<n$. If $U_k=\varnothing$, then the $k$th term never attains the maximum in \eqref{eq:polytope_expression} and can therefore be removed without changing $p$. Suppose instead that $U_k\neq\varnothing$. Since $U_k$ is not full dimensional, the system \eqref{eq:monomial-region} contains an implicit equality. Thus, for some $j\neq k$ with $\va_j\neq\va_k$,
$$
\langle \va_j, \vx \rangle + c_j = \langle \va_k, \vx \rangle + c_k
$$
for every $\vx\in U_k$. Hence, whenever the $k$th term attains the maximum, the $j$th term attains it as well. Removing the $k$th monomial therefore leaves the represented function unchanged.

Conversely, suppose that the $k$th monomial can be removed without changing $p$, and assume for contradiction that $\dim U_k=n$. Then the system \eqref{eq:monomial-region} has no implicit equalities, so
there exists $\bar{\vx}\in U_k$ at which all of its inequalities are strict \cite[e.g.,][\S 8.1]{schrijver1998theory}. Thus,
$$
    \langle \va_j, \bar{\vx} \rangle + c_j < \langle \va_k, \bar{\vx} \rangle + c_k \qquad \text{for every } j \neq k.
$$
The $k$th monomial is therefore the unique maximizer at $\bar{\vx}$, and removing it necessarily changes the value of $p$ at $\bar{\vx}$---a contradiction.
\end{proof}

This characterization establishes a direct correspondence between
non-redundant monomials and linear regions of a signomial.

\begin{proposition} \label[proposition]{prop:monomials-linear-regions}
    Let $p$ be a signomial. The number of non-redundant monomials that appear in the expression of $p$ is the same as the number of linear regions of $p$.
\end{proposition}

\begin{proof}
After removing redundant monomials, we can write
$$
p(\vx) = c_1 \odot \vx^{\odot\va_1} \oplus \dotsb \oplus c_k \odot \vx^{\odot \va_k}.
$$
For each monomial, let $U_i$ denote its associated polyhedron. By \cref{lem:redundant-monomial}, each $U_i$ is full-dimensional. Moreover, $U_i$ is a convex polyhedron, and hence the closure of its connected interior and $p$ is represented on $U_i$ by the affine map
$$
\vx \mapsto c_i + \langle \va_i, \vx \rangle.
$$
It remains only to verify maximality.

Suppose that $U_i$ is strictly contained in the closure $T$ of a connected open set on which $p$ is affine. Since $T$ contains the full-dimensional set $U_i$, the affine representation of $p$ on $T$ must be
$$
p(\vx) = c_i + \langle \va_i, \vx \rangle.
$$
Since the polyhedra $U_1, \ldots, U_k$ cover $\RR^n$ and $T$ strictly contains $U_i$, the interior of $T$ must intersect the interior of some $U_j$ for $j \not= i$. At any point $\vx$ in this intersection, the $j$th monomial is the unique maximizer, so 
$$
p(\vx) = c_j + \langle \va_j, \vx \rangle > c_i + \langle \va_i, \vx \rangle,
$$
which contradicts the assumption of an affine representation of $p$ in $T$.  Thus, each $U_i$ is a linear region of $p$.

Conversely, every linear region of $p$ contains a full-dimensional subset on which one of its monomials attains the maximum, and therefore intersects with the interior of the corresponding $U_i$. By maximality, the two must coincide. So the linear regions of $p$ are in one-to-one correspondence with its non-redundant monomials.
\end{proof}

For neural networks that admit a signomial representation (rather than a rational signomial representation), this correspondence allows the number of linear regions to be read directly from the tropical expression.

\begin{corollary}
    Suppose $f$ is a scalar-valued neural network whose weights are nonnegative and activation functions are signomials with non-negative exponents. Let $f_\mathrm{trop}$ be a signomial representing $f$ (this exists by \cref{rmk:nonneg-exponents}).
    Then the number of linear regions of $f$ is equal to the number of non-redundant monomials of $f_\mathrm{trop}$.
\end{corollary}

\begin{proof}
    This follows directly from \cref{prop:monomials-linear-regions}.
\end{proof}

The characterization in \cref{lem:redundant-monomial} also gives a practical pruning procedure. As discussed previously, whether a polyhedron is full-dimensional can be determined using linear programming. We can therefore test each monomial of a signomial and remove those whose associated polyhedra are not full-dimensional, yielding an equivalent expression containing no redundant monomials.

For rational signomials, however, redundancy is more subtle. Pruning the numerator and denominator independently need not produce a minimal representation of the rational signomial, since additional simplifications may arise through common structure between the two. For example,
$$
f(x,y) = (0\oplus x\oplus y\oplus x\odot y) \oslash(0\oplus x) = (0\oplus y)\oslash 0.
$$
The numerator and denominator of the first expression contain no redundant monomials when considered separately; however, the rational expression as a whole admits a strictly simpler representation. Developing methods for identifying such globally minimal representations is an interesting direction for future work.

%% file: arXiv/subsec_lr_new.tex
We now extend the ideas of \cref{subsec:linear-regions-signomials} to the computation of linear regions of rational signomials. We first give an algorithm for an arbitrary rational signomial (\cref{alg:linear-regions}) and prove its correctness. We then exploit the compositional structure of neural networks to obtain a layerwise variant (\cref{alg:layerwise-linear-regions}). To the best of our knowledge, these are the first algorithms that compute exact descriptions of linear regions, in the sense of \cref{def:linear-region}, for arbitrary input and output dimensions and general rational signomial activations.

\begin{figure}[ht]
\centering
\def\cliprad{3.5}%
\def\shiftp{0.5}%
\pgfmathsetmacro{\px}{\shiftp}%
\pgfmathsetmacro{\py}{\shiftp}%
\def\shiftq{1.5}%
\pgfmathsetmacro{\sx}{-\shiftq}%
\pgfmathsetmacro{\sy}{-\shiftq}%

\begin{subfigure}[t]{0.32\textwidth}
\centering
\begin{tikzpicture}[scale=0.45]
  \begin{scope}
    \clip (-\cliprad,-\cliprad) rectangle (\cliprad,\cliprad);
    \draw[thick, darkblue] (\px,\py) -- (-\cliprad,\py);
    \draw[thick, darkblue] (\px,\py) -- (\px,-\cliprad);
    \draw[thick, darkblue] (\px,\py) -- ({\px+\cliprad},{\py+\cliprad});
  \end{scope}
  \node at ({\px-1.8},{\py+1.8}) {$U_1$};
  \node at ({\px+1.8},{\py-0.5}) {$U_2$};
  \node at ({\px-1.8},{\py-1.8}) {$U_3$};
  \draw[thin, gray] (-\cliprad,-\cliprad) rectangle (\cliprad,\cliprad);
\end{tikzpicture}
\caption{Linear regions of $p$.}
\label{fig:linear-regions-p}
\end{subfigure}
\hfill
\begin{subfigure}[t]{0.32\textwidth}
\centering
\begin{tikzpicture}[scale=0.45]
  \begin{scope}
    \clip (-\cliprad,-\cliprad) rectangle (\cliprad,\cliprad);
    \draw[thick, darkred] (\sx,\sy) -- (\cliprad,\sy);
    \draw[thick, darkred] (\sx,\sy) -- (\sx,\cliprad);
    \draw[thick, darkred] (\sx,\sy) -- ({\sx-\cliprad},{\sy-\cliprad});
  \end{scope}
  \node at ({\sx+2},{\sy+2}) {$V_1$};
  \node at ({\sx-1.5},{\sy+0.5}) {$V_2$};
  \node at ({\sx+2},{\sy-1.5}) {$V_3$};
  \draw[thin, gray] (-\cliprad,-\cliprad) rectangle (\cliprad,\cliprad);
\end{tikzpicture}
\caption{Linear regions of $q$.}
\label{fig:linear-regions-q}
\end{subfigure}
\hfill
\begin{subfigure}[t]{0.32\textwidth}
\centering
\begin{tikzpicture}[scale=0.45]
  \begin{scope}
    \clip (-\cliprad,-\cliprad) rectangle (\cliprad,\cliprad);
    \draw[thick, darkblue] (\px,\py) -- (-\cliprad,\py);
    \draw[thick, darkblue] (\px,\py) -- (\px,-\cliprad);
    \draw[thick, darkblue] (\px,\py) -- ({\px+\cliprad},{\py+\cliprad});
    \draw[thick, darkred] (\sx,\sy) -- (\cliprad,\sy);
    \draw[thick, darkred] (\sx,\sy) -- (\sx,\cliprad);
    \draw[thick, darkred] (\sx,\sy) -- ({\sx-\cliprad},{\sy-\cliprad});
  \end{scope}
  \draw[thin, gray] (-\cliprad,-\cliprad) rectangle (\cliprad,\cliprad);
\end{tikzpicture}
\caption{$p \oslash q$ is linear on all intersections.}
\label{fig:linear-regions-intersection-sub}
\end{subfigure}
\caption{Linear regions of $p$, $q$, and the polyhedral subdivision associated to $p \oslash q$.}
\label{fig:linear-regions-intersection}
\end{figure}

Let $\vf = \vp \oslash \vq$ and let $U_1, \ldots, U_s$ and $V_1, \ldots, V_t$ denote the linear regions of $\vp$ and $\vq$, respectively. Since $\vf = \vp - \vq$, it is affine on every intersection $U_i \cap V_j$; these intersections therefore form a polyhedral subdivision of the input space on which $\vf$ is piecewise affine (see \cref{fig:linear-regions-intersection}). They need not,
however, coincide with the linear regions of $\vf$: some intersections may be lower-dimensional, while adjacent intersections carrying the same affine map may combine to form a single linear region. \cref{alg:linear-regions} accounts for precisely these two phenomena.

\begin{notation}
    If $\vf : \RR^n \to \RR^m$ is a piecewise linear function and $U \subset \RR^n$ a subset on which $\vf$ is represented by an affine map, we write $L(\vf, U)$ for the affine map representing $\vf$ on $U$.
\end{notation}

\begin{algorithm}[H]
  \caption{Linear regions of rational signomials}
    \begin{algorithmic}[1]
        \Require Vectors of signomials $\vp = (p_1, \dotsc, p_m), \vq =  (q_1, \dotsc, q_m)$ in $n$ variables.
        \State Compute the linear regions $U_1, \dotsc, U_\ell$ of $\vp$, and set $L_i = L(\vp, U_i)$.
        \State Compute the linear regions $V_1, \dotsc, V_r$ of $\vq$, and set $S_j = L(\vq, V_j)$.
        \State Compute the pairs $(i, j)$ such that $U_i \cap V_j$ has dimension $n$
        \For{$(i, j)$ such that $U_i \cap V_j$ has dimension $n$}
            \State Compute the affine map $T_{ij} = L_i - S_j$
        \EndFor
        \State Set $S$ to be the set of all $T_{ij}$
        \For{$T \in S $}
            \State Compute the set $G(T)$ of indices $(i, j)$ such that $\dim U_i \cap V_j = n$ and $T = T_{ij}$.
            \State Compute the set $C(T)$ of connected full-dimensional components of
            $$
                \bigcup_{(i, j) \in G(T)} U_i \cap V_j
            $$
        \EndFor
        \State \Return $\bigcup_{T \in S} C(T)$.
    \end{algorithmic}
    \label{alg:linear-regions}
\end{algorithm}

The only step of \cref{alg:linear-regions} that requires some further explanation is the computation of the full-dimensional components $C(T)$. For a fixed affine map $T$, consider the full-dimensional polyhedra $U_i \cap V_j$ on which $\vf$ is represented by $T$. Two such polyhedra belong to the same full-dimensional component precisely when they can be connected through a sequence of polyhedra meeting along sets of codimension at most one. Thus, $C(T)$ can be computed as the connected components of the graph whose vertices are these polyhedra, with an edge whenever two polyhedra intersect in dimension at least $n-1$. This construction is illustrated in \cref{fig:gluing-linear-regions}.

\begin{figure}[ht]
\centering
\definecolor{regionpink}{RGB}{238,165,196}
\definecolor{regionteal}{RGB}{125,207,196}
\definecolor{regionpurple}{RGB}{195,179,234}
\def\R{3}%
\begin{subfigure}[b]{0.48\textwidth}
\centering
\begin{tikzpicture}[scale=0.8]
  \begin{scope}
    \clip (-\R,-\R) rectangle (\R,\R);
    \fill[regionpurple] (-\R,0) -- (0,0) -- (0,\R) -- (-\R,\R) -- cycle;

    \fill[regionpurple] (0,0) -- (\R,-\R) -- (\R,\R) -- (0,\R) -- cycle;
    \fill[regionteal] (-\R,-\R) -- (-\R,0) -- (0,0) -- (\R,-\R) -- cycle;
    \draw[thick] (-\R,0) -- (0,0) -- (\R,-\R);

    \draw[very thick, dashed, regionpink!60!black] (0,0) -- (0,\R);
  \end{scope}
  \node[regionpurple!45!black] at (-1.55,1.9) {\footnotesize $U_k\cap V_\ell$};
  \node[regionpurple!45!black] at (1.5,1.9) {\footnotesize $U_i\cap V_j$};
  \draw[thin, gray] (-\R,-\R) rectangle (\R,\R);
\end{tikzpicture}
\caption{$f$ is represented by the \emph{same} affine map on $U_i\cap V_j$ and $U_k\cap V_\ell$}
\label{fig:gluing-linear-regions-before}
\end{subfigure}
\hfill
\begin{subfigure}[b]{0.48\textwidth}
\centering
\begin{tikzpicture}[scale=0.8]
  \begin{scope}
    \clip (-\R,-\R) rectangle (\R,\R);
    \fill[regionpurple] (-\R,0) -- (0,0) -- (\R,-\R) -- (\R,\R) -- (-\R,\R) -- cycle;
    \fill[regionteal] (-\R,-\R) -- (-\R,0) -- (0,0) -- (\R,-\R) -- cycle;
    \draw[thick] (-\R,0) -- (0,0) -- (\R,-\R);
  \end{scope}
  \node[regionpurple!45!black] at (0.2,1.9) {$R$};
  \draw[thin, gray] (-\R,-\R) rectangle (\R,\R);
\end{tikzpicture}
\caption{The two regions glue into a non-convex linear region $R$.}
\label{fig:gluing-linear-regions-after}
\end{subfigure}
\caption{Intersections assembling into a linear region.}
\label{fig:gluing-linear-regions}
\end{figure}

This gluing can be determined entirely from the intersections of the polyhedra. For each $T$, form a graph whose vertices are the full-dimensional polyhedra $U_i \cap V_j$ associated with $T$, with an edge between two vertices whenever the corresponding polyhedra intersect in dimension at least $n-1$. The full-dimensional components are precisely the unions corresponding to connected components of this graph. The following straightforward geometric observation justifies this construction.

\begin{lemma}\label[lemma]{lem:glue-full-dimension}
    Let $P$ and $Q$ be $n$-dimensional polyhedra in $\mathbb R ^n$. Suppose that $P \cap Q$ has dimension at least $n-1$. Then the union $P \cup Q$ is the closure of a connected open set. Conversely, if $P \cap Q$ has dimension strictly less than $n-1$ then $P \cup Q$ is not the closure of a connected open set.
\end{lemma}

\begin{proof}
    Suppose that $\dim (P \cap Q) = n$. Then we have $\mathring P \cap \mathring Q \ne \varnothing$ so $\mathring P \cup \mathring Q$ is a connected open set whose closure is $P \cup Q$. Next, assume that $\dim (P \cap Q) = n-1$. 
    Let $H$ be the affine hull of $P \cap Q$. By assumption, $H$ is a hyperplane, $P \cap H$ is a facet of $P$, and $Q \cap H$ is a facet of $Q$.

    Let $S = \operatorname{relint}_H(P \cap Q)$ and define $U := \mathring{P} \cup S \cup \mathring{Q}$. We will show that $U$ is open and connected. Since the closure of $U$ is $P \cup Q$, the conclusion will follow. 

    For the openness of $U$, it suffices to show that every element of $U$ has an open neighborhood contained in $U$. Since this is clearly the case for elements in $\mathring P$ and $\mathring Q$, we can restrict to elements that lie in $S$.

    Let $\vn$ be a vector that is normal to $H$, so that 
    $$
    H = \{ \vx \mid \langle \vn, \vx \rangle = b\} 
    $$
    for some constant $b$.
    Since $P \cap H$ is a facet of $P$, it must be the case that (up to replacing $\vn$ by $-\vn$) every element $\vx \in P$ satisfies $\langle \vn, \vx \rangle \ge b$. Since $\dim(P \cap Q) = n - 1$, $P$ and $Q$ must lie on opposite sides of $H$, otherwise their interiors would intersect in a neighborhood of a point in $S$, which would imply that $\dim(P \cap Q) = n$. Hence, every $\vx \in Q$ satisfies $\langle \vn, \vx \rangle \le b$.

    Let $\vx \in S = \operatorname{relint}_H(P \cap Q)$ relative to $H$. Then $\vx$ does not lie in any other facet of $P$, so there exists an open neighborhood $V$ of $\vx$ that lies in the intersection of all the open half-spaces corresponding to other facets of $P$. Similarly, up to shrinking $V$, we can assume that it also lies in the intersection of all the open half-spaces corresponding to other facets of $Q$.

For $\vy \in V$, if $\langle \vn, \vy \rangle > b$, then $\vy$ strictly satisfies the inequality corresponding to $H$, as well as all the remaining inequalities defining $P$, and thus $\vy \in \mathring P$. Similarly, if $\langle \vn, \vy \rangle < b$, then $\vy \in \mathring Q$.

It remains to consider the case $\langle \vn, \vy \rangle = b$, so that $\vy \in H$. By the choice of $V$, the point $\vy$ satisfies strictly
all the remaining facet inequalities of both $P$ and $Q$. Hence $\vy \in \operatorname{relint}_H(P \cap Q) = S$. Therefore, in all three cases,
$$
\vy \in \mathring P \cup S \cup \mathring Q = U.
$$
Thus $V \subseteq U$, which proves that $U$ is open.

To see that $U$ is connected, observe that every point of $S$ lies in the closure of both $\mathring P$ and $\mathring Q$. Hence $\mathring P \cup S$ and $S \cup \mathring Q$ are connected. Since these two sets intersect in $S$, their union $U = \mathring P \cup S \cup \mathring Q$ is connected. This proves the first implication.

    Now for the converse, suppose that $P \cap Q$ has dimension less than $n-1$. If $P \cap Q$ is empty then it is clear that $P \cup Q$ is disconnected, and thus cannot be the closure of a connected open set. Now assume that $P \cap Q \ne \varnothing$.
    Let $F$ (resp $G$) be a facet of $P$ (resp $Q$) containing $P \cap Q$. Since $\codim(P \cap Q) > 1$, we can choose $F$ and $G$ to have distinct supporting hyperplanes. Let $\vu$ be a vector normal to $F$ such that all elements $ \vx \in P$ satisfy $\langle \vx, \vu\rangle \le b$ for some $b$.
    Similarly, let $\vv$ be a vector normal to $G$ such that all elements $\vx \in Q$ satisfy $\langle \vx, \vv\rangle \le c$ for some $c$. Set $R$ to be the polyhedron defined by inequalities
    \begin{align*}
    \langle \vx, \vu\rangle & \ge b \\
     \langle \vx, \vv\rangle & \ge c
    \end{align*}
    Since the supporting hyperplanes of $F$ and $G$ are distinct, it must be the case that $\dim R = n$.
    In particular, $\mathring R$ is open in $\RR^n$. Now suppose that $P \cup Q$ is the closure of a connected open set $U$.
    Let $V$ be the interior of $P \cup Q$. 
    Note that $P \cap Q$ doesn't intersect with $V$, as $P \cap Q$ lies in the closure of $\mathring R$, and $\mathring R$ doesn't intersect with $P \cup Q$. This implies that $V = \mathring P \cup \mathring Q$, which is disconnected. Since $\overline U = P \cup Q$, $U$ must intersect with both connected components of $V$. Since $U$ is connected and $U \subset V$, this is a contradiction. 
\end{proof}

It follows that the connected components of the graph above give exactly the full-dimensional components required in \cref{alg:linear-regions}. In particular, all of the operations in the algorithm reduce to computations with polyhedra; we discuss their
practical implementation using linear programming below.

We can now establish the correctness of the algorithm.

\begin{theorem}[Correctness of \cref{alg:linear-regions}] \label{thm:linear-region-algorithm}
    \cref{alg:linear-regions} computes the linear regions of a vector of rational signomials $\vf = \vp \oslash \vq$.
\end{theorem}

\begin{proof}
    Let $U_1, \dotsc, U_\ell$ be the linear regions of $\vp$ and $L_i = L(\vp, U_i)$. Similarly, let $V_1, \dotsc, V_r$ be the linear regions of $\vq$, and set $S_j = L(\vq, V_j)$.
    We take $T_{ij} = L_i - S_j$ and set
    $$
        S = \left\{ T_{ij} \mid U_i \cap V_j \text{ has dimension } n  \right\},
    $$
    and set $A$ to be the set of all intersections $U_i \cap V_j$ of dimension $n$.
    By \cref{lem:polyhedral-covering}, this collection of polyhedra covers $\RR^n$, and intersections can only happen on boundaries.
    For $T \in S$, let $G(T)$ be the graph whose vertices are pairs $(i, j)$ such that $\dim U_i \cap V_j = n$ and $T = T_{ij}$, with an edge between $(i, j)$ and $(k, h)$ whenever $(U_i \cap V_j) \cap (U_k \cap V_h)$ has dimension at least $n-1$, and let
    $$
        D(T) = \bigcup_{(i, j) \in G(T)} U_i \cap V_j.
    $$
    Notice that $\vf$ coincides with $T$ on $D(T)$, and in fact we have
    $$
        D(T) = \left\{ \mathbf x \in \RR^n \mid \text{There exists a connected full-dimensional set } N \text{ such that } x \in N \text{ and } \vf_{\mid_N} = T\right\}.
    $$
    In particular all linear regions of $\vf$ associated with the affine map $T$ are subsets of $D(T)$. To determine the linear regions of $\vf$ corresponding to $T$, it suffices to compute the full-dimensional components of $D(T)$.
    Note that by \cref{lem:glue-full-dimension}, the full-dimensional components of $D(T)$ are precisely unions of the form
    $$
        \bigcup_{(i, j) \in C} U_i \cap V_j
    $$
    where $C$ is a connected component of the graph $G(T)$.
    By construction, $\vf$ is represented by the affine map $T$ on any such component, and it is easy to check that each of these is maximal, i.e., that it is \emph{not} strictly contained in the closure of a connected open set $U$ such that $\vf$ is affine on $\overline U$.
    This shows that every element of the output of \cref{alg:linear-regions} is a linear region. To see that the algorithm enumerates every linear region, note that by \cref{lem:polyhedral-covering}, the elements of the set $A$ cover all of $\RR^n$ and can only intersect on codimension at least 1 components. This implies that regions in the output of \cref{alg:linear-regions} must also cover all of $\RR^n$. In particular, any linear region of $\vf$ must intersect with the interior of one of the regions in the output of the algorithm, which implies that they coincide.
\end{proof}

\begin{remark}
If, instead, connected components are computed after including appropriate lower-dimensional cells and intersections, the same construction yields the closed connected linear regions considered by
\cite{stargalla2025computational}.
\end{remark}

\begin{remark}\label[remark]{rmk:complexity-linear-regions}
The computational cost of \cref{alg:linear-regions} depends strongly on the number of monomials in the numerator and denominator. For a single rational signomial with $s$ numerator monomials and $t$
denominator monomials, the algorithm requires at most $(st)^2$ polyhedral non-emptiness or dimension checks. This makes the simplification of tropical expressions, discussed in \cref{subsec:linear-regions-signomials}, particularly important in practice.
\end{remark}

Neural networks possess additional compositional structure that can be exploited computationally. By \cref{thm:nn-rational-signomial-form}, a network can be represented as a composition
$$
\vf = \vf_d \circ \ldots \circ \vf_1
$$
of rational signomial maps. Rather than first expanding this composition into a single rational signomial and then applying \cref{alg:linear-regions}, we can compute its polyhedral subdivision incrementally, one layer at a time. This yields the following layerwise variant of \cref{alg:linear-regions}.

\begin{algorithm}
  \caption{Layerwise linear regions of a composition}
    \begin{algorithmic}[1]
        \Require Vectors of rational signomials $\vf_k:\RR^{n_{k-1}}\to\RR^{n_k}$ for $k=1,\dotsc,d$.
        \State Set $\mathcal P_0=\{(\RR^{n_0},\id_{\RR^{n_0}})\}$. \Comment{Start with the whole space}
        \For{$k=1,\dotsc,d$}
            \State Set $\mathcal P_k=\varnothing$. \Comment{Contains the polyhedral subdivision of the composition up to $k$}
            \For{$(P,L)\in\mathcal P_{k-1}$}
                \State Compute the polyhedral subdivision $\mathcal C$ associated to the rational signomial $\vf_k \circ L$.
                \State Compute the set $\mathcal D$ of full dimensional intersections of elements of $\mathcal C$ with $P$.
                \For{each full-dimensional polyhedron $R\in\mathcal D$}
                    \State Add $(R,L(\vf_k\circ L,R))$ to $\mathcal P_k$.
                \EndFor
            \EndFor
        \EndFor
        \State Set $S_d$ to be the set of affine maps appearing in $\mathcal P_d$.
        \For{$T\in S_d$}
            \State Compute the set $I_d(T)$ of polyhedra $R$ such that $(R,T)\in\mathcal P_d$.
            \State Compute the set $C_d(T)$ of full-dimensional components of
            $$
                \bigcup_{R\in I_d(T)} R.
            $$
        \EndFor
        \State \Return $\{U\mid U \in C_d(T) \text{ for some } T \in S_d\}$.
    \end{algorithmic}
    \label{alg:layerwise-linear-regions}
\end{algorithm}

\begin{theorem} \label{thm:linear-region-composition-algorithm}
    \cref{alg:layerwise-linear-regions} computes the linear regions of a composition of rational signomials $\vf = \vf_d\circ\dots\circ\vf_1$.
\end{theorem}

\begin{proof}
    First, note that for each $k$, the polyhedra in $\mathcal P_k$ form a subdivision of $\RR^{n_0}$, since for each $k$, the polyhedra in $\mathcal P_k$ are obtained by subdividing the polyhedra in $\mathcal P_{k-1}$, which themselves form a subdivision of $\RR^{n_0}$.

    Moreover by construction, for each $k$, the composition $\vf_k \circ \dotsb \circ \vf_1$ can be represented by an affine map on each polyhedron in $\mathcal P_k$.
    Thus, the full composition $\vf$ is linear on each polyhedron in the final subdivision $\mathcal P_d$. Just as in the proof of \cref{thm:linear-region-algorithm}, to obtain the linear regions of $\vf$, it suffices to take unions of the polyhedra in $\mathcal P_d$ that correspond to the same affine map and then compute the full-dimensional components of these unions.
\end{proof}

\cref{alg:layerwise-linear-regions} can be viewed as a generalization of the approach of \cite{wang2022estimation} to arbitrary rational signomial maps. An important distinction is that recovering linear regions in the sense of \cref{def:linear-region} requires identifying polyhedra that meet along sets of codimension at most one, rather than merely testing whether their intersection is non-empty; see also \cite{stargalla2025computational}.

\paragraph{Polyhedral Computations.} Both algorithms rely on a small number of basic operations on polyhedra: testing feasibility, checking full dimensionality, and determining whether intersections have dimension at least $n-1$. Each can be reduced to linear programming, as we briefly describe below.

\paragraph{Checking Non-Emptiness.} Given an arbitrary polyhedron $P = P(\mA, \vb)$, we need to check whether $P = \varnothing$, i.e., whether $P$ is \emph{feasible}.
This can be achieved by solving the following linear program:
\begin{align*}
    \text{minimize } & s \\
    \text{subject to } & \va_i^{\top} \vx \le \vb_i + s, i=1, \dotsc, m \\
    & s \ge 0
\end{align*}
There is always a solution $s^*$ to this problem. Moreover, it is straightforward to check that $s^* = 0$ if and only if $P$ is feasible.

\paragraph{Checking Full Dimensionality.} A polyhedron $P = \{ \vx \in \RR^n \mid \mA \vx \le \vb \}$ is full dimensional if and only if it contains a ball of positive radius. 
To check this, we first discard every \emph{trivial row}, that is, every row $j$ with $\va_j = \mathbf 0 $ and $b_j \ge 0$. We then solve the linear problem
\begin{align*}
    \text{maximize } & \varepsilon \\
    \text{subject to } & \mA \vx + \varepsilon \cdot \mathbf{1} \le \vb
\end{align*}
Note that if $P$ is non-empty and $\varepsilon$ is the solution then $\varepsilon \ge 0$.
If $\varepsilon > 0$ or if the problem is unbounded then $P$ is full dimensional. Otherwise, $\dim P < n$.

\paragraph{Computing Dimension.} The computation for the dimension of a polyhedron can be approached as follows: 
\begin{enumerate}
    \item First, check whether the polyhedron is empty and check whether it is full dimensional.
    \item If not, then solve the $m$ linear programs
    \begin{align*}
    \text{maximize } & s = b_i - \va_i^{\top} \vx  \\
    \text{subject to } & \mA \vx \le \vb
    \end{align*}
    \item For a given $i$, if the solution $s^* = 0$ then the inequality $\va_i^{\top} \vx \le b_i$ is in fact an implicit equality. If $s^* > 0$ or if the problem is unbounded then the inequality is not an implicit equality. Let $\mA^=$ be the submatrix of $\mA$ whose rows correspond to the $\va_i$s that yield implicit equalities.
    \item The dimension of $P$ is then $n - \mathrm{rank}\left(\mA^=\right)$.
\end{enumerate}

%% file: arXiv/library.tex
\section{\textsc{TropicalNN.jl}: A Symbolic Computation Framework for Neural Networks} \label{sec:library}

We developed \textsc{TropicalNN.jl}, an extensible Julia library for
symbolic tropical and polyhedral computations with neural networks.  Built on top of the OSCAR computer algebra system \citep{OSCAR}, our
library provides a computational realization of the framework developed in this paper and, more broadly, a collection of tools for manipulating neural networks as tropical algebraic objects.

The functionality of \textsc{TropicalNN.jl} is organized around four
main components: 
\begin{enumerate}[(i)]
\item Our library implements symbolic algebra for
signomials and rational signomials, including arithmetic and composition;
\item It also converts neural networks with tropical activations into rational signomial representations and provides tools for simplifying these expressions by pruning redundant monomials;
\item It further implements the algorithms of \cref{sec:expressivity} for computing exact linear regions of rational signomials and neural networks; and finally,
\item It provides tools for computing neural Hoffman constants, together with lower and upper bounds that can be used when exact computation becomes expensive. 
\end{enumerate}

Together, these components provide a unified pipeline from a neural network architecture to its symbolic tropical representation and its subsequent algebraic and geometric analysis.

\textsc{TropicalNN.jl} interfaces with the HiGHS linear programming solver \citep{huangfu2018parallelizing} and Polymake \citep{gawrilow2000polymake} for polyhedral computations, allowing
symbolic, optimization, and polyhedral methods to be combined within a single computational framework. In this section, we describe each of the four components and illustrate their use with examples.

\subsection{Symbolic Signomial Algebra}

A core component of \textsc{TropicalNN.jl} is a symbolic algebra for signomials and rational signomials. Given signomials $p$ and $q$, the library supports the basic tropical operations $p\oplus q$, $p\odot q$, and $p\oslash q$, as well as exponentiation and
composition. These operations return symbolic signomial or rational
signomial objects, allowing for more complicated tropical expressions to be constructed and manipulated within the library. In
particular, composition is essential for representing the successive
layers of a neural network symbolically.

The library allows the numerical type of the coefficients and exponents to be specified by the user. In particular, computations can be performed using floating-point arithmetic, as is natural when working with the parameters of a trained neural network, or using exact rational arithmetic when exact symbolic computation is desired.

Signomial algebra differs in important ways from classical polynomial
algebra as signomial exponents may be arbitrary real
numbers rather than integers and the tropical operations admit
simplifications that have no direct analogue in ordinary polynomial
arithmetic. One important example is exponentiation by a nonnegative
real number $r$. For a signomial, we use the tropical identity
$$
    \left(\bigoplus_i c_i \vx^{\va_i} \right)^{\odot r} = \bigoplus_i c_i ^{\odot r} \vx^{\odot r \va_i},
$$
which follows from the \emph{Freshman's dream} identity $(a \oplus b)^{\odot r} = a ^{\odot r} \oplus b ^{\odot r}$, which always holds \citep[e.g.,][]{maclagan2015introduction}. Even when $r$ is a natural
number, this differs from ordinary polynomial exponentiation, which
generally introduces additional terms. Exploiting the tropical identity therefore avoids unnecessary growth in the number of monomials and is particularly important for maintaining tractable symbolic representations as computations are composed across the layers of a neural network.

\cref{fig:signomial-setup} illustrates the construction of signomials
and several of the algebraic operations implemented in
\textsc{TropicalNN.jl}. In the Julia implementation, these tropical operations are represented through the standard arithmetic syntax, as illustrated in \cref{fig:signomial-setup}.

\begin{figure}[H]
\begin{lstlisting}
# a = max(1.5x, 0.5y, x + 0.25y)
a = Signomial([0, 0, 0], [[1.5, 0.0], [0.0, 0.5], [1.0, 0.25]])
# b = max(0.5x + 0.5y, 1.5y)
b = Signomial([0, 0], [[0.5, 0.5], [0.0, 1.5]])

a + b   # max(1.5x, 0.5y, x+0.25y, 0.5x+0.5y, 1.5y)
a * b   # max(2y, 0.5x+y, x+1.75y, 1.5x+0.75y, 1.5x+1.5y, 2x+0.5y)

# max(x, y)
f1 = Signomial([0, 0], [[1.0, 0.0], [0.0, 1.0]])
# max(2x, 0.5y)
f2 = Signomial([0, 0], [[2.0, 0.0], [0.0, 0.5]])
# max(x+y, 2x)
g  = Signomial([0, 0], [[1.0, 1.0], [2.0, 0.0]])

# The composition: max(3x, 2x+y, x+0.5y, 2x, 2y, 1.5y)
comp(g, [f1, f2])
\end{lstlisting}
\caption{Algebraic operations on signomials.}
\label{fig:signomial-setup}
\end{figure}

\subsection{Tropical Representations of Neural Networks}

Another central functionality of \textsc{TropicalNN.jl} is the conversion of neural networks into symbolic tropical representations. Given a neural network with tropical activation functions, the library constructs a rational signomial representing the function computed by the network. The resulting object can then be manipulated using the signomial algebraic rules described above and passed directly to the other algorithms implemented by the library, allowing for the computation of linear regions and neural Hoffman constants.

The construction follows the correspondence between neural networks and rational signomials established in \cref{thm:nn-rational-signomial-form}, which \textsc{TropicalNN.jl} implements by translating each layer into a rational signomial map and composing the resulting expressions.

\paragraph{Computing Tropical Representations.} Given a neural network $\mathcal{N}$ with weight matrices $\mW_0, \dotsc, \mW_m$ and bias vectors $\vb_0, \dotsc, \vb_m$, write $\vx^{\odot A}$ for the vector whose $r$th entry is $x_1^{\odot A_{r1}} \odot \dotsb \odot x_n^{\odot A_{rn}}$. Let $\sigma_i = \vp_i \oslash \vq_i$ be the activation function for the $i$th layer. All tropical operations on vectors are taken componentwise. Then an affine transformation can be written tropically as
\begin{align*}
    \mW_i \vx + \vb_i & = \vb_i \odot \vx^{\odot \mW_i}
\end{align*}
so the $i$th layer has the rational signomial representation
\begin{equation} \label{eqn:one-layer-tropical}
\sigma_i(\mW_i \vx + \vb_i) = (\vp_i \oslash \vq_i)\left(\vb_i \odot \vx^{\odot \mW_i}\right).
\end{equation}

Applying this construction successively to each layer yields a symbolic tropical representation of the complete network; see
\cref{alg:tropical-form}.

\begin{algorithm}
    \begin{algorithmic}[1]
        \Require Neural network $\mathcal{N}$ with weight matrices $\mW_0, \dotsc, \mW_m$ and bias vectors $\vb_0, \dotsc, \vb_m$.
        \For{$i=0, \dotsc, m-1$}
            \State Compute the tropical form $\mF_i(\vx)$ of $\sigma_i(\mW_i\vx + \vb_i)$ using \cref{eqn:one-layer-tropical}.
        \EndFor
        \State Compute the tropical form $\mF_m(\vx)$ of $\mW_m \vx + \vb_m$.
        \State Compute the composition $\mathcal{T}(\mathcal{N})$ of rational signomials $\mF_m \circ \dotsb \circ \mF_0$.
        \State \Return $\mathcal{T}(\mathcal{N})$
    \end{algorithmic}
    \caption{Computation of the rational signomial of a neural network}
    \label{alg:tropical-form}
\end{algorithm}

In the library, this entire construction is represented through the
\texttt{tropicalize} operation. Thus, a network represented using the
native neural-network objects of \textsc{TropicalNN.jl} can be converted directly into a symbolic rational signomial, as shown in \cref{fig:mlp-to-trop}.

\begin{figure}[H]
\begin{lstlisting}
W1 = Rational{BigInt}[0 1; 1 0]
b1 = Rational{BigInt}[1, 1]

W2 = Rational{BigInt}[1 -1]
b2 = Rational{BigInt}[0]

network = NeuralNetwork(
    AffineLayer(W1, b1),
    ActivationLayer(relu(), 2),
    AffineLayer(W2, b2)
)

# F is the tropical representation of the network
F = tropicalize(network)
# F[1] = (max(0, 1 + x_2)) / (max(0, 1 + x_1))
\end{lstlisting}
\caption{Constructing a two-layer ReLU MLP and computing its tropical representation.}
\label{fig:mlp-to-trop}
\end{figure}

\paragraph{Simplifying Tropical Representations.} Explicitly composing the rational signomial maps associated with successive layers can lead to expressions containing many monomials, some of which are redundant. This is important computationally: the
complexity of subsequent symbolic and polyhedral operations, including the computation of linear regions, grows rapidly with the number of monomials in the representation; see \cref{rmk:complexity-linear-regions}. Simplification is therefore an important part of the symbolic pipeline implemented in \textsc{TropicalNN.jl}.

The library provides a \texttt{prune} operation implementing the
geometric characterization of redundant monomials established in
\cref{lem:redundant-monomial}. For each monomial of a signomial, we
compute the polyhedron on which that monomial attains the maximum. By
\cref{lem:redundant-monomial}, the monomial is redundant precisely when this polyhedron is not full-dimensional, yielding the pruning procedure in \cref{alg:pruned-expression}.

\begin{algorithm}[H]
  \caption{Pruning tropical expressions.}
    \begin{algorithmic}[1]
        \Require Signomial $g$ in $n$ variables.
        \For{each monomial $c_i \odot \vx^{\odot \va_i}$ of $g$}
            \State Compute the corresponding polyhedron $P_i$.
            \If{$P_i$ has dimension less than $n$}
                \State Discard the $i$th monomial
            \EndIf
        \EndFor
        \State \Return g
    \end{algorithmic}
    \label{alg:pruned-expression}
\end{algorithm}

For a rational signomial, \textsc{TropicalNN.jl} applies this procedure independently to the numerator and denominator. As discussed in \cref{subsec:monomials}, this removes all monomials that are redundant within the individual signomial expressions, although it does not in general produce a globally minimal representation of the rational signomial.

\begin{figure}[H]
\begin{lstlisting}
f_big = Signomial([0, 0, 0, 0, 0, 0], [[1, 0], [0, 1], [2, 2], [1, 1], [2, 0], [0, 2]])
f_pruned = prune(f_big)
println(f_pruned)
# max(x_2, 2x_2, x_1, 2x_1, 2x_1 + 2x_2)
\end{lstlisting}
\caption{Pruning \texttt{f\_big} to eliminate redundant monomials.}
\label{fig:pruning}
\end{figure}

\subsection{Exact Computation of Linear Regions}

\textsc{TropicalNN.jl} provides tools for computing the exact 
linear regions of rational signomials and neural networks as explicit polyhedral subsets of the input space, which can then be studied and manipulated using the polyhedral tools provided by OSCAR and Polymake.

The library supports two computational strategies, which correspond to the algorithms developed in \cref{sec:expressivity}. The first operates directly on the representation of the network as a composition, applying \cref{alg:layerwise-linear-regions} successively across its
layers. The second converts the network into a rational signomial using \texttt{tropicalize}, simplifies the resulting expression using \texttt{prune}, and then applies \cref{alg:linear-regions}. The second procedure is summarized in the following pipeline

$$
\mathcal{N} \xrightarrow{\texttt{tropicalize}} p \oslash q \xrightarrow{\texttt{prune}} p'\oslash q' \xrightarrow{\texttt{linear\_regions}} \mathcal{S},
$$
where $\mathcal{S}$ is the exact collection of linear regions of
$\mathcal{N}$.

Both algorithms are implemented through the \texttt{linear\_regions} function in \textsc{TropicalNN.jl}. The same function can also be applied directly to a rational signomial, in which case the library computes its regions using \cref{alg:linear-regions}. For example, \cref{fig:linear-regions-rat} constructs a rational signomial and enumerates its linear regions.

\begin{figure}[H]
\begin{lstlisting}
# p = max(x, y, x+y)
num_exps   = [[1, 0], [0, 1], [1, 1]]
num_coeffs = [0, 0, 0]
p = Signomial(num_coeffs, num_exps)

# q = max(x, x+2y)
den_exps   = [[1, 0], [1, 2]]
den_coeffs = [0, 0]
q = Signomial(den_coeffs, den_exps)

f = p / q

# f has 4 linear regions
regions_f = linear_regions(f; mode = HiGHSMode())
\end{lstlisting}
\caption{Enumerating the linear regions of the rational signomial $\mathtt{f} = p/q$, using the HiGHS-based solver and \cref{alg:linear-regions}.}
\label{fig:linear-regions-rat}
\end{figure}

For a neural network, \texttt{linear\_regions} can instead operate directly on the layerwise representation, avoiding the need to construct the complete tropical expression explicitly. For the network introduced in \cref{fig:mlp-to-trop}, this requires only the following call.

\begin{figure}[H]
\begin{lstlisting}
regions_nn = linear_regions(network; mode = HiGHSMode())
\end{lstlisting}
\caption{Enumerating the linear regions of the two-layer MLP from
\cref{fig:mlp-to-trop} layer by layer using \cref{alg:layerwise-linear-regions}.}
\label{fig:linear-regions-mlp}
\end{figure}

\subsection{Computing Hoffman Constants for Neural Network Linear Regions}\label{sec:pena_alg}

\textsc{TropicalNN.jl} provides tools for computing Hoffman constants associated with rational signomials and neural networks. As discussed in \cref{sec:expressivity}, the Hoffman constant provides geometric information about the distribution of linear regions in the input space: in particular, it controls the distance from an arbitrary input point to the furthest linear region. The library makes this geometric quantity directly obtainable from the symbolic tropical representation of a network.

\paragraph{Computation.} The library implements the Pe\~na--Vera--Zuluaga (PVZ) algorithm of \cite{pena2018algorithm} for exact computation of Hoffman constants. Given a rational signomial, \texttt{hoffman\_constant} constructs the linear systems associated with its linear regions and applies the PVZ
algorithm to compute their Hoffman constants. These are then combined to obtain the Hoffman constant of the complete tropical expression, following the construction given in \cref{sec:expressivity}.

It turns out that exact computation can become expensive as the size of the underlying linear systems increases. In particular, the PVZ algorithm requires a combinatorial search over subsets of constraints that involves solving many linear programs. To support computations for larger expressions, \textsc{TropicalNN.jl} also provides \texttt{lower\_hoffman\_constant} and \texttt{upper\_hoffman\_constant}, which compute rigorous lower and upper bounds using the approximation methods developed in
\cref{sec:expressivity}. These methods provide certified information about the Hoffman constant
without requiring its exact computation. In all these implementations, linear programs are solved numerically using the GLPK solver.

The exact and bounding computations are exposed through a common
interface, as illustrated in \cref{fig:hoffman}.

\begin{figure}[H]
\begin{lstlisting}
# f = p/q defined above
hoff_exact = hoffman_constant(f)       # exact value of the Hoffman constant
hoff_upper = upper_hoffman_constant(f) # upper bound for the Hoffman constant
hoff_lower = lower_hoffman_constant(f) # lower bound for the Hoffman constant
\end{lstlisting}
\caption{Computing exact and bounding estimates of the Hoffman constant for the rational signomial map $\mathtt{f} = p/q$, using the PVZ algorithm.}
\label{fig:hoffman}
\end{figure}

The same functionality can be applied to the tropical representations of neural networks produced by \texttt{tropicalize}, allowing Hoffman constants and their bounds to be incorporated directly into the symbolic analysis pipeline provided by \textsc{TropicalNN.jl}.

%% file: arXiv/experiments.tex
\section{Computational Examples} \label{sec:experiments}

In this section, we demonstrate the capabilities of \textsc{TropicalNN.jl} through a series of proof-of-concept examples illustrating the symbolic analyses enabled by our framework. Unlike conventional empirical evaluations, our goal is not to benchmark predictive performance, but rather to showcase the types of exact algebraic and geometric computations that become possible once neural networks are represented as tropical algebraic objects.

Symbolic computation is inherently more computationally demanding than standard floating point arithmetic, and exact methods are therefore most tractable for relatively small network architectures and low-dimensional input spaces. Existing approaches to linear-region analysis similarly focus primarily on relatively small networks (e.g., \cite{serra2018bounding} or \cite{piwek2023exact}).  Our framework computes explicit polyhedral descriptions of linear regions and enables additional analyses including monomial complexity and Hoffman-based geometric bounds.

We illustrate the functionalities of \textsc{TropicalNN.jl} in four directions. First, we study the monomial structure of tropical neural-network representations and
its relationship with linear regions. Second, we use exact polyhedral representations to study how the geometry of linear regions evolves during training. Third, we investigate the computational reach of the framework across network architectures and input dimensions. Finally, we demonstrate the Hoffman-based bounds from \cref{sec:theory-hoffman-computation} and use them to construct bounded domains guaranteed to
intersect every linear region.

The code used in these experiments can be found in the repositories linked in \cref{tab:links}.

\begin{table}[ht]
    \centering
    \begin{tabular}{|l|l|}
        \hline
       \textsc{TropicalNN.jl}  & \href{https://github.com/Paul-Lez/TropicalNN.jl}{\texttt{github.com/Paul-Lez/TropicalNN.jl}} \\
       \hline
       Experiments  & \href{https://github.com/Paul-Lez/tropicalnn}{\texttt{github.com/Paul-Lez/tropicalnn}} \\
       \hline
    \end{tabular}
    \caption{Links to the repositories used for the experiments}
    \label{tab:links}
\end{table}

\subsection{Neural Network Monomials}
The tropical representation of a neural network naturally introduces a new measure of algebraic complexity through its monomials. 
Unlike linear regions, this quantity has received little attention in the literature. In this subsection, we briefly explore its properties empirically.
We first examine the relationship between monomials and linear regions, then investigate the prevalence of redundant monomials in tropical representations, and finally illustrate how (non-redundant) monomial counts capture well-known trends in neural network expressivity.

\begin{figure}[H]
    \centering
    \begin{subfigure}[t]{0.48\textwidth}
        \centering
        \begin{tikzpicture}
        \begin{axis}[
            xlabel={Monomials in $p, q$},
            ylabel={Linear Regions of $p \oslash q$},
            xmin=0, xmax=1000,
            ymin=0, ymax=3000,
            xtick={100, 500, 1000},
            ytick={500, 2000, 3000},
            legend pos=north west,
            ymajorgrids=true,
            grid style=dashed,
            width=\textwidth,
            height=5cm,
        ]
        \addplot[
            color=blue,
            mark=square,
            restrict expr to domain={\thisrow{NVars}}{3:3},
            ] table [
                col sep=comma,
                x=NMonomials,
                y=MeanNumRegions
            ] {data/monomial-experiment/signomials_regions.csv};
        \addlegendentry{3 variables}
        \addplot[
            color=red,
            mark=triangle,
            restrict expr to domain={\thisrow{NVars}}{4:4},
            ] table [
                col sep=comma,
                x=NMonomials,
                y=MeanNumRegions
            ] {data/monomial-experiment/signomials_regions.csv};
        \addlegendentry{4 variables}
        \end{axis}
        \end{tikzpicture}
    \caption{Evolution of the number of linear regions in terms of the number of monomials}
        \label{fig:linear_regions_monomials}
    \end{subfigure}
    \hfill
    \begin{subfigure}[t]{0.48\textwidth}
        \centering
        \begin{tikzpicture}
            \begin{axis}[
                xlabel={Network Width},
                ylabel={Rate of Pruning},
                xmin=1.5, xmax=8.5,
                ymin=0, ymax=1, 
                xtick={2,3,4,5,6,7,8},
                ytick={0, 0.2, 0.4, 0.6, 0.8, 1.0},
                legend pos=north west,
                grid=both,
                grid style={line width=.1pt, draw=gray!20},
                major grid style={line width=.2pt,draw=gray!50},
                width=\textwidth,
                height=5cm,
                mark size=2.5pt
            ]
            \addplot[
                color=blue,
                mark=square*,
                thick
            ] table [
                col sep=comma,
                x=Width,
                y=Rate_Init
            ] {data/monomial-experiment/rate_of_pruning.csv};
            \addlegendentry{Untrained}
            \addplot[
                color=red,
                mark=triangle*,
                thick
            ] table [
                col sep=comma,
                x=Width,
                y=Rate_Trained
            ] {data/monomial-experiment/rate_of_pruning.csv};
            \addlegendentry{Trained}
            \end{axis}
        \end{tikzpicture}
        \caption{Proportion of redundant monomials as a function of the width}
        \label{fig:pruning_rate}
    \end{subfigure}
    \caption{In \cref{fig:linear_regions_monomials}, we consider the number of linear regions and monomials of a rational signomial in 3 variables (blue) and 4 variables (red).
    In \cref{fig:pruning_rate} we consider one-hidden-layer neural networks of widths $\{2,3,4,5,6,7,8\}$ with two-dimensional input spaces and one-dimensional output spaces.
    We measure the ratio of the number of monomials before and after pruning (rate of pruning) when the neural network is randomly initialized (blue) and when it has been trained on a classification problem (red).
    Rates of pruning are averaged across thirty random initializations.}
\end{figure}

\paragraph{Comparison to Linear Regions.} Monomials are the terms in the tropical expressions of neural networks. Although connected to linear regions, their enumerations are in general distinct. \cref{fig:linear_regions_monomials} shows the evolution of the average number of linear regions over 30 samples as we vary the number of monomials of randomly generated rational signomials in 3 and 4 variables.

\begin{figure}[H]
    \centering
    \begin{subfigure}[t]{0.48\textwidth}
        \centering
        \begin{tikzpicture}
            \begin{axis}[
                ymode=log,
                xlabel={Network Width Parameter $w$},
                ylabel={Monomial Count},
                xmin=1.5, xmax=8.5,
                xtick={2,3,4,5,6,7,8},
                legend pos=north west,
                grid=both,
                grid style={line width=.1pt, draw=gray!20},
                major grid style={line width=.2pt,draw=gray!50},
                width=\textwidth,
                height=5cm,
                mark size=2.5pt
            ]
            \addplot[
                color=green!60!black,
                mark=*,
                thick,
            ] table [
                col sep=comma,
                x=Width,
                y=Unpruned_Avg
            ] {data/width-depth-monomials/sweep_onelayer.csv};
            \addlegendentry{1-Layer}
            \addplot[
                color=purple,
                mark=diamond*,
                thick,
            ] table [
                col sep=comma,
                x=Depth, 
                y=Unpruned_Avg
            ] {data/width-depth-monomials/sweep_twolayer.csv};
            \addlegendentry{2-Layer}
            \end{axis}
        \end{tikzpicture}
        \caption{Number of monomials as a function of the width.}
        \label{fig:monomial_expressivity_unpruned}
    \end{subfigure}
    \hfill
    \begin{subfigure}[t]{0.48\textwidth}
        \centering
        \begin{tikzpicture}
            \begin{axis}[
                xlabel={Network Width Parameter $w$},
                ylabel={Monomial Count},
                xmin=1.5, xmax=8.5,
                xtick={2,3,4,5,6,7,8},
                legend pos=north west,
                grid=both,
                grid style={line width=.1pt, draw=gray!20},
                major grid style={line width=.2pt,draw=gray!50},
                width=\textwidth,
                height=5cm,
                mark size=2.5pt
            ]
            \addplot[
                color=green!60!black,
                mark=*,
                thick,
            ] table [
                col sep=comma,
                x=Width,
                y=Pruned_Avg
            ] {data/width-depth-monomials/sweep_onelayer.csv};
            \addlegendentry{1-Layer}
            \addplot[
                color=purple,
                mark=diamond*,
                thick,
            ] table [
                col sep=comma,
                x=Depth, 
                y=Pruned_Avg
            ] {data/width-depth-monomials/sweep_twolayer.csv};
            \addlegendentry{2-Layer}
            \end{axis}
        \end{tikzpicture}
        \caption{Number of pruned monomials as a function of the width.}
        \label{fig:monomial_expressivity_pruned}
    \end{subfigure}

    \caption{In \cref{fig:monomial_expressivity_unpruned} and \cref{fig:monomial_expressivity_pruned}, we measure the number of unpruned and pruned monomials for one-hidden-layer and two-hidden-layer neural networks, with two-dimensional input spaces and one-dimensional output spaces. The one-hidden-layer neural networks have architecture $[2,w + 2,1]$ and the two-hidden-layer neural networks have architecture $[2,w,2,1]$, where $w$ is the width parameter. Here, the architectures are chosen so that for a fixed $w$, both have the same number of neurons.}
    \label{fig:combined_master_figure}
\end{figure}

\paragraph{Monomial Redundancy.} When directly computing the tropical representation of a neural network using \cref{alg:tropical-form}, we note that many of the monomials are redundant. The pruning ratio (i.e., the proportion of redundant terms among all monomials) is explored for neural networks of different widths in \cref{fig:pruning_rate}.  Interestingly, at this scale, the pruning ratio does not vary significantly between trained and untrained neural networks.

\paragraph{Measuring Expressivity.} Considering monomials as a measure of the algebraic complexity of tropical representations of neural networks (after pruning), we compare a few different architectures in \cref{fig:monomial_expressivity_pruned}. The corresponding unpruned monomial counts are shown in \cref{fig:monomial_expressivity_unpruned}.

In the examples considered, increasing the width of a layer in a deeper network produces rapid growth in the number of monomials in the resulting tropical expression, both before and, to a lesser extent, after pruning. Moreover, for a fixed total number of neurons, the deeper architectures considered here produce representations with substantially more non-redundant monomials. Although monomial count is representation-dependent and therefore should not be interpreted as an
intrinsic measure of network expressivity, this behavior mirrors the well-established increase in expressive power associated with network depth \citep{telgarsky2016benefits}.

\subsection{Geometry of Linear Regions During Training}

The explicit polyhedral representations outputted by \textsc{TropicalNN.jl} allow us to study substantially more than the number of linear regions. To illustrate this, we track the tropical and polyhedral geometry of a neural network throughout training. At each checkpoint, we compute the network's tropical representation and its complete collection of linear regions, allowing us to monitor quantities such as monomial complexity, region volume, combinatorial complexity, and Hoffman constants alongside conventional training metrics.

\begin{figure}[H]
    \centering
    \includegraphics[width=0.7\textwidth]{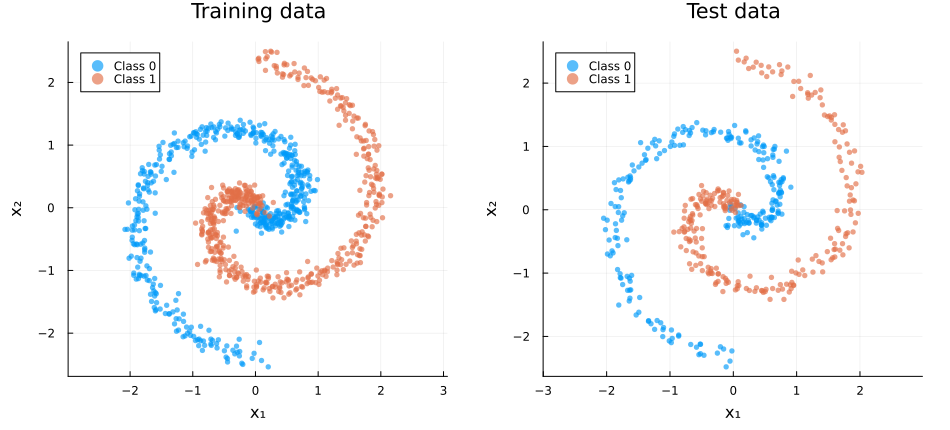}
    \caption{Training and held-out data for the binary classification task used in the experiment.}
    \label{fig:area_dynamics_data}
\end{figure}

\paragraph{Experimental Setup.} We train a one-hidden-layer neural network with two-dimensional input, a hidden layer of width 10, and a scalar output on a binary classification task for 200 epochs. The two-dimensional input space allows us to compute and analyze the complete polyhedral decomposition induced by the network throughout training. At each epoch, we compute the tropical representation of the network, prune its redundant monomials, and enumerate its linear regions exactly using \textsc{TropicalNN.jl}. Additionally, we record the number of monomials before and after pruning, the volumes of the bounded linear regions, the maximum number of vertices of the regions, and the
Hoffman constant associated with the tropical representation of the network. These quantities provide complementary algebraic and geometric descriptions
of how the piecewise-linear function represented by the network and the corresponding linear regions evolve during training.

\Cref{fig:training_dynamics} shows the conventional learning dynamics for this experiment. We use these curves primarily as a reference against which to compare the evolution of the tropical and polyhedral quantities displayed below.

\begin{figure}[H]
    \centering
    \begin{minipage}[t]{0.44\textwidth}
        \centering
        \begin{tikzpicture}
            \begin{axis}[
                width=\textwidth,
                height=5cm,
                xlabel={Epoch},
                ylabel={Accuracy},
                grid=major,
                legend pos=north west,
                legend cell align={left},
                legend style={font=\small},
                table/col sep=comma
            ]
            \addplot[thick, blue, line width=1.25pt] 
                table [x=Epoch, y=Train_Accuracy] {data/training-measurements/training_metrics.csv};
            \addlegendentry{Train}
            
            \addplot[thick, red, dashed, line width=1.25pt] 
                table [x=Epoch, y=Validation_Accuracy] {data/training-measurements/training_metrics.csv};
            \addlegendentry{Validation}
            \end{axis}
        \end{tikzpicture}
    \end{minipage}
    \hfill
    \begin{minipage}[t]{0.44\textwidth}
        \centering
        \begin{tikzpicture}
            \begin{axis}[
                width=\textwidth,
                height=5cm,
                xlabel={Epoch},
                ylabel={Loss},
                grid=major,
                legend pos=north east,
                legend cell align={left},
                legend style={font=\small},
                table/col sep=comma
            ]
            \addplot[thick, blue, line width=1.25pt]
                table [x=Epoch, y=Train_Loss] {data/training-measurements/training_metrics.csv};
            \addlegendentry{Train}

            \addplot[thick, red, dashed, line width=1.25pt]
                table [x=Epoch, y=Validation_Loss] {data/training-measurements/training_metrics.csv};
            \addlegendentry{Validation}
            \end{axis}
        \end{tikzpicture}
    \end{minipage}
    \caption{
    Here we train a one-hidden-layer neural network of width 10 on a classification task.
    The network is trained with binary cross-entropy using Adam with learning rate $10^{-3}$, batch size 16, and 200 epochs.
    On the left, we monitor its training and validation accuracy; on the right, we monitor its training and validation loss.
    }
    \label{fig:training_dynamics}
\end{figure}

\paragraph{Algebraic and Volumetric Evolution.}
\Cref{fig:volume_dynamics} shows that training changes both the algebraic representation of the network and the geometry of its linear regions. The number of monomials in the tropical representation changes over the course of training. Notably, the gap between the counts before and after pruning is considerable throughout training. This shows that the tropical expression of the network obtained by composing all layers contains a substantial number of redundant terms throughout training.

\begin{figure}[H]
    \centering
    \begin{minipage}[t]{0.44\textwidth}
        \centering
        \begin{tikzpicture}
            \pgfplotsset{set layers}
            \begin{axis}[
                width=0.86\textwidth,
                height=4.3cm,
                xlabel={Epoch},
                ylabel={Mean Finite Area},
                ymode=log,
                axis y line*=left,
                grid=major,
                legend style={at={(0.5,1.05)}, anchor=south, font=\small},
                legend columns=2,
                legend cell align={left},
                table/col sep=comma
            ]
            \addplot[thick, darkgray, line width=1.25pt]
                table [x=Epoch, y=Mean_Finite_Region_Volume] {data/training-measurements/whole_plane_region_stats.csv};
            \addlegendentry{Mean}
            \addlegendimage{thick, orange, dashed, line width=1.25pt}
            \addlegendentry{Median}
            \end{axis}
            \begin{axis}[
                width=0.86\textwidth,
                height=4.3cm,
                ylabel={Median Finite Area},
                axis y line*=right,
                axis x line=none,
                table/col sep=comma
            ]
            \addplot[thick, orange, dashed, line width=1.25pt]
                table [x=Epoch, y=Median_Finite_Region_Volume] {data/training-measurements/whole_plane_region_stats.csv};
            \end{axis}
        \end{tikzpicture}
    \end{minipage}
    \hfill
    \begin{minipage}[t]{0.44\textwidth}
        \centering
        \begin{tikzpicture}
            \begin{axis}[
                width=\textwidth,
                height=4.3cm,
                xlabel={Epoch},
                ylabel={Monomial Count},
                grid=major,
                legend cell align={left},
                legend style={at={(0.5,1.05)}, anchor=south, font=\small},
                table/col sep=comma
            ]
            \addplot[thick, blue, const plot, line width=1.25pt]
                table [x=Epoch, y=Pre_Pruning] {data/training-measurements/monomial_counts.csv};
            \addlegendentry{Before pruning}

            \addplot[thick, red, dashed, const plot, line width=1.25pt]
                table [x=Epoch, y=Post_Pruning] {data/training-measurements/monomial_counts.csv};
            \addlegendentry{After pruning}
            \end{axis}
        \end{tikzpicture}
    \end{minipage}
    \caption{
    For the same network and training run as in \cref{fig:training_dynamics}, we monitor the mean and median area of its bounded linear regions on the left, and its monomial counts before and after pruning on the right.
    }
    \label{fig:volume_dynamics}
\end{figure}

We also note that certain geometric quantities associated with bounded linear regions also evolve significantly. Notably, the mean area varies over several orders of magnitude during training, and occasionally spikes, which is due to certain polyhedra becoming very large, whereas the median remains comparatively small and substantially more stable.
This discrepancy indicates a highly heterogeneous distribution of region sizes: most bounded regions are comparatively small, and 
changes in the mean are driven by a relatively small number of large bounded regions.
We note that exact enumeration of the regions makes this distinction directly accessible, which would not be visible from the mere linear region count.

\paragraph{Combinatorial Complexity and Hoffman Constants.} In contrast, a different perspective appears in the combinatorial complexity of the
regions. As we see in \cref{fig:geometric_training_dynamics}, the maximum number of vertices/edges of an individual region changes only slightly over the course of training. Thus, in this example, substantial changes in region volume and tropical representation occur without a corresponding increase in the maximal combinatorial complexity of the individual polyhedra.

The Hoffman constant associated with the tropical representation, however, varies substantially during training. Since it enters our bound on the distance from a point in the input space to the farthest linear region, it therefore provides geometric information about the polyhedral systems defining the network's
linear regions. Its evolution during training shows significant variation in this aspect of the network geometry that is not captured by the combinatorial complexity of the individual regions. Together, these experiments demonstrate that training can affect distinct aspects of the piecewise-linear function in different ways, through its algebraic representation; the sizes and combinatorial structure of its linear regions; and the geometric quantities controlling their location
in the input space.

As mentioned previously, the purpose of this experiment is not to draw general conclusions about the geometry of neural-network training from a single small architecture. Rather, it demonstrates the type of analysis made possible by exact symbolic calculations of the linear regions of a neural network: quantities that describe the complete collection of these regions can be computed and tracked alongside conventional learning metrics, which provides a computational framework for studying aspects of neural-network geometry that are invisible to linear-region counts alone.

\begin{figure}[H]
    \centering
    \begin{minipage}[t]{0.44\textwidth}
        \centering
        \begin{tikzpicture}
            \begin{axis}[
                width=\textwidth,
                height=4.3cm,
                xlabel={Epoch},
                ylabel={Maximum Count},
                ymin=4.8,
                ymax=8.2,
                ytick={5,6,7,8},
                grid=major,
                legend cell align={left},
                legend style={at={(0.5,1.05)}, anchor=south, font=\small},
                table/col sep=comma
            ]
            \addplot[thick, darkgreen, const plot, line width=1.25pt]
                table [x=Epoch, y=Max_Vertices_Per_Region] {data/training-measurements/polyhedra_stats.csv};
            \addlegendentry{Max. vertices}
            \addplot[thick, orange, dashed, const plot, line width=1.25pt]
                table [x=Epoch, y=Max_Facets_Per_Constituent_Polyhedron] {data/training-measurements/polyhedra_stats.csv};
            \addlegendentry{Max. edges}
            \end{axis}
        \end{tikzpicture}
    \end{minipage}
    \hfill
    \begin{minipage}[t]{0.44\textwidth}
        \centering
        \begin{tikzpicture}
            \begin{axis}[
                width=\textwidth,
                height=4.3cm,
                xlabel={Epoch},
                ylabel={Hoffman Constant},
                ymode=log,
                grid=major,
                table/col sep=comma
            ]
            \addplot[thick, purple, line width=1.25pt]
                table [x=Epoch, y=Hoffman_Constant] {data/training-measurements/hoffman_constants.csv};
            \end{axis}
        \end{tikzpicture}
    \end{minipage}
    \caption{
    Maximum vertex count and the Hoffman constant.
    }
    \label{fig:geometric_training_dynamics}
\end{figure}

\subsection{Architecture and Scale}

Having examined how the tropical and polyhedral structure of a fixed network evolves during training, we now study how our algorithms behave across different network architectures and problem dimensions. 
We begin by varying the width and depth of randomly initialized networks to study how the architecture affects the number of linear regions and the computational cost of their exact enumeration. We then consider networks trained on MNIST to study how these methods behave in higher dimensional input space. These experiments illustrate complementary aspects of our computational framework, namely its ability to compare the exact piecewise-linear complexity of different neural network architectures as well as its applicability beyond small input spaces and synthetic data.

\paragraph{Width and Depth.} We first study how the number of linear regions and the cost of their exact enumeration vary with the network architecture. We consider networks of increasing width and depth, using both ReLU and maxout activations. For the width experiment, we used architectures with one hidden layer of the form $[2, X, 1]$ where $X \in \{10, 20, \ldots, 60\}$. For the depth experiment, we used architectures of the form $[2, 10, \dotsc, 10, 1]$ with $1 \le d \le 6$ hidden layers of width 10. For maxout networks, each maxout unit takes 3 inputs. For each architecture and activation, we generate 30 randomly initialized networks and compute their linear regions exactly. The average numbers of linear regions and computation times are reported in \cref{fig:width-linear-regions} and \cref{fig:depth-linear-regions}.

\newcommand{\plotwidthdepthseries}[6]{%
    \addplot[#1, filter discard warning=false,
        error bars/.cd,
            y dir=both,
            y explicit,
        /pgfplots/.cd,
        x filter/.code={%
            \ifnum\pdfstrcmp{\thisrow{Network}}{#5}=0\relax
                \ifnum\pdfstrcmp{\thisrow{Sweep}}{#6}=0\relax
                \else
                    \def\pgfmathresult{}%
                \fi
            \else
                \def\pgfmathresult{}%
            \fi
        }
    ] table [x=#2, y=#3, y error=#4, col sep=comma] {data/width-depth-linear-regions/linear_regions.csv};%
}

As we see in \cref{fig:width-linear-regions}, increasing the width of a single hidden layer results in a significant increase in the number of linear regions. ReLU and maxout networks have quite similar behavior across the range of our experiment, both in the number of regions and in the time required for their enumeration. The computational cost increases together with the number of regions, reaching tens of seconds for networks of width 60, for which approximately 1,800 regions are enumerated on average.

\begin{figure}[H]
    \centering
    \begin{subfigure}[t]{0.48\textwidth}
        \centering
        \begin{tikzpicture}
            \begin{axis}[
                width=\textwidth, height=5cm,
                xlabel={Width}, ylabel={Number of regions},
                xtick={10,20,30,40,50,60}, grid=major,
                legend pos=north west
            ]
                \plotwidthdepthseries{blue, mark=square*, thick}{Width}{MeanNumRegions}{StdNumRegions}{relu}{width}
                \addlegendentry{ReLU}
                \plotwidthdepthseries{red, mark=triangle*, thick}{Width}{MeanNumRegions}{StdNumRegions}{maxout}{width}
                \addlegendentry{Maxout}
            \end{axis}
        \end{tikzpicture}
    \end{subfigure}
    \hfill
    \begin{subfigure}[t]{0.48\textwidth}
        \centering
        \begin{tikzpicture}
            \begin{axis}[
                width=\textwidth, height=5cm,
                xlabel={Width}, ylabel={Time (seconds)},
                xtick={10,20,30,40,50,60}, grid=major,
                legend pos=north west
            ]
                \plotwidthdepthseries{blue, mark=square*, thick}{Width}{MeanTimeSeconds}{StdTimeSeconds}{relu}{width}
                \addlegendentry{ReLU}
                \plotwidthdepthseries{red, mark=triangle*, thick}{Width}{MeanTimeSeconds}{StdTimeSeconds}{maxout}{width}
                \addlegendentry{Maxout}
            \end{axis}
        \end{tikzpicture}
    \end{subfigure}
    \caption{Linear-region counts and enumeration times as the width of a one-hidden-layer network varies; error bars show one standard deviation over 30 random initializations.}
    \label{fig:width-linear-regions}
\end{figure}

\begin{figure}[H]
    \centering
    \begin{subfigure}[t]{0.48\textwidth}
        \centering
        \begin{tikzpicture}
            \begin{axis}[
                width=\textwidth, height=5cm,
                xlabel={Number of hidden layers}, ylabel={Number of regions},
                xtick={1,2,3,4,5,6}, grid=major,
                legend pos=north west
            ]
                \plotwidthdepthseries{blue, mark=square*, thick}{HiddenLayers}{MeanNumRegions}{StdNumRegions}{relu}{depth}
                \addlegendentry{ReLU}
                \plotwidthdepthseries{red, mark=triangle*, thick}{HiddenLayers}{MeanNumRegions}{StdNumRegions}{maxout}{depth}
                \addlegendentry{Maxout}
            \end{axis}
        \end{tikzpicture}
    \end{subfigure}
    \hfill
    \begin{subfigure}[t]{0.48\textwidth}
        \centering
        \begin{tikzpicture}
            \begin{axis}[
                width=\textwidth, height=5cm,
                xlabel={Number of hidden layers}, ylabel={Time (seconds)},
                xtick={1,2,3,4,5,6}, grid=major,
                legend pos=north west
            ]
                \plotwidthdepthseries{blue, mark=square*, thick}{HiddenLayers}{MeanTimeSeconds}{StdTimeSeconds}{relu}{depth}
                \addlegendentry{ReLU}
                \plotwidthdepthseries{red, mark=triangle*, thick}{HiddenLayers}{MeanTimeSeconds}{StdTimeSeconds}{maxout}{depth}
                \addlegendentry{Maxout}
            \end{axis}
        \end{tikzpicture}
    \end{subfigure}
    \caption{Linear-region counts and enumeration times as the depth of a width-$10$ network varies; error bars show one standard deviation over 30 random initializations.}
    \label{fig:depth-linear-regions}
\end{figure}

Similarly, increasing depth also results in a significant growth in the number of linear regions (\cref{fig:depth-linear-regions}). ReLU and maxout networks also behave similarly here for depths up to five hidden layers, while at six hidden layers the maxout networks show a larger average number of regions. Enumeration time follows the same overall trend, which increases substantially with depth as the resulting polyhedral decomposition becomes more complex. The error bars show one standard deviation over 30 random initializations; they tend to grow across networks as the architectures become larger, showing increased variability, especially for the deeper networks. The growing error bars with depth suggest that networks with the same architecture can produce polyhedral decompositions with quite different complexity depending on initialization.

\paragraph{MNIST.} We next consider a higher-dimensional setting by applying our exact linear-region computation to neural networks trained on the MNIST dataset. We train ReLU classifiers with architectures of the form $[784, w, 10]$ and $[784, w, w, 10]$, for $w \in \{4, 5, 6, 7, 8\}$. For each architecture, we work with 30 samples and compute the exact linear regions of the trained network (after removing the softmax on the output layer) and report the resulting average region counts, test accuracies, and computation times in \cref{fig:mnist-linear-regions}.

\newcommand{\plotmnistseries}[4]{%
    \addplot[#1, filter discard warning=false,
        error bars/.cd,
            y dir=both,
            y explicit,
        /pgfplots/.cd,
        x filter/.code={%
            \ifnum\pdfstrcmp{\thisrow{Activation}}{relu}=0\relax
                \ifnum\thisrow{HiddenLayers}=#3\relax
                \else
                    \def\pgfmathresult{}%
                \fi
            \else
                \def\pgfmathresult{}%
            \fi
        }
    ] table [x=Width, #2, col sep=comma] {#4};%
}

\begin{figure}
    \centering
    \begin{subfigure}[t]{0.32\textwidth}
        \centering
        \begin{tikzpicture}
            \begin{axis}[
                width=\textwidth, height=5cm,
                xlabel={Width}, ylabel={Number of regions},
                xtick={4,5,6,7,8}, grid=major, scaled y ticks=false,
                legend pos=north west, legend style={font=\scriptsize}
            ]
                \plotmnistseries{blue, mark=square*, thick}{y=MeanNumRegions, y error=StdNumRegions}{1}{data/mnist/linear_regions.csv}
                \addlegendentry{1 layer}
                \plotmnistseries{red, mark=triangle*, thick}{y=MeanNumRegions, y error=StdNumRegions}{2}{data/mnist/linear_regions.csv}
                \addlegendentry{2 layers}
            \end{axis}
        \end{tikzpicture}
    \end{subfigure}
    \hfill
    \begin{subfigure}[t]{0.32\textwidth}
        \centering
        \begin{tikzpicture}
            \begin{axis}[
                width=\textwidth, height=5cm,
                xlabel={Width}, ylabel={Test accuracy (\%)},
                xtick={4,5,6,7,8}, grid=major,
                legend pos=south east, legend style={font=\scriptsize}
            ]
                \plotmnistseries{blue, mark=square*, thick}{y expr=100*\thisrow{MeanTestAccuracy}, y error expr=100*\thisrow{StdTestAccuracy}}{1}{data/mnist/metrics.csv}
                \addlegendentry{1 layer}
                \plotmnistseries{red, mark=triangle*, thick}{y expr=100*\thisrow{MeanTestAccuracy}, y error expr=100*\thisrow{StdTestAccuracy}}{2}{data/mnist/metrics.csv}
                \addlegendentry{2 layers}
            \end{axis}
        \end{tikzpicture}
    \end{subfigure}
    \hfill
    \begin{subfigure}[t]{0.32\textwidth}
        \centering
        \begin{tikzpicture}
            \begin{axis}[
                width=\textwidth, height=5cm,
                xlabel={Width}, ylabel={Time (seconds)},
                xtick={4,5,6,7,8}, grid=major,
                legend pos=north west, legend style={font=\scriptsize}
            ]
                \plotmnistseries{blue, mark=square*, thick}{y=MeanTimeSeconds, y error=StdTimeSeconds}{1}{data/mnist/linear_regions.csv}
                \addlegendentry{1 layer}
                \plotmnistseries{red, mark=triangle*, thick}{y=MeanTimeSeconds, y error=StdTimeSeconds}{2}{data/mnist/linear_regions.csv}
                \addlegendentry{2 layers}
            \end{axis}
        \end{tikzpicture}
    \end{subfigure}
    \caption{MNIST ReLU-network test accuracy and exact linear-region enumeration as the width varies. Error bars show one standard deviation over 30 random initializations.}
    \label{fig:mnist-linear-regions}
\end{figure}

The behavior in the number of linear regions is substantially different for the one- and two-hidden-layer architectures. For a single hidden layer, the number of regions grows comparatively moderately over the range of widths we studied. Adding a second hidden layer leads to sharp growth: at width 8, the two-hidden layer network architecture yields more than $10^4$ linear regions on average. 
This yields another concrete demonstration of the substantial increase in piecewise-linear complexity that can arise with greater depth.

An interesting observation is that such a difference in geometric complexity is not reflected by an equally large difference in predictive performance. Test accuracy improves with width for both architectures, but the one- and two-hidden-layer networks achieve similar accuracies for the larger widths considered, despite their substantially different numbers of linear regions. 
This emphasizes that linear-region count measures the complexity of the piecewise-linear function represented by the network, rather than serving as an explicit indicator for predictive accuracy. The variability over trained networks also differs between these quantities: in particular, the two-hidden-layer networks show significant variation at larger widths. Their test accuracies show less variability, however, which provides some evidence that networks with similar predictive performance can still differ quite significantly in the geometric complexity of the functions they represent.

The computation times also demonstrate the cost of exact symbolic enumeration. While the one-hidden-layer networks remain comparatively cheap over this range, the cost for two-hidden-layer networks increases significantly with the width, reaching the order of $10^3$ seconds at width 8. 
Thus, while these exact polyhedral computations remain possible in a 784-dimensional input space, the computation becomes highly challenging as the architectural complexity increases.

\subsection{Bounding the Search Region for Neural Network Linear Regions}

Finally, we evaluate the computational tools that we developed for bounding the domain required to hit all linear regions of a neural network. We first compare the exact PVZ computation of the Hoffman constant with brute-force enumeration and with the lower- and upper-bound approximations introduced in \cref{alg:lower_hoffman,alg:upper_hoffman}. We then use these quantities to construct an effective bounded search domain for the linear regions of a neural network.

\paragraph{Computing and Bounding Hoffman Constants.} We evaluate the different methods on randomly generated rational signomials. We generate signomials $p$ and $q$ with $s$ and $t$ monomials, respectively, by sampling exponent matrices $A \in \RR^{s \times n}$ and $B \in \RR^{t \times n}$ with entries sampled uniformly from $[0,1]$. From these expressions, we construct the matrix
in \eqref{eq:def-hoff-rational-map} and compute its Hoffman constant using both brute-force enumeration and the PVZ algorithm, together with the lower and upper bounds of \cref{alg:lower_hoffman,alg:upper_hoffman}. For the randomized lower bound, we sample one tenth of the number of candidate subsets evaluated by the brute-force approach.

For each configuration of $(s, t, n)$, we average the results over 30 randomly generated examples. The mean absolute errors relative to the brute-force exact Hoffman constant and the mean computation times are given in \cref{tab:hoffman_constants} for the different values of $s$, $t$, and $n$.

\begin{table}[ht]
\centering
\caption{Mean absolute errors relative to the exact Hoffman constant computed using brute-force enumeration and mean computation times (seconds), averaged over 30 randomly generated samples.}
\label{tab:hoffman_constants}
\resizebox{\textwidth}{!}{%
\begin{filecontents*}{data/hoffman-signomials/hoffman_summary.csv}
Benchmark,MP,MQ,N,CoefficientPolicy,LowerSamplePolicy,NumSamples,MeanCellMatrices,StdCellMatrices,MeanBruteForceCandidates,StdBruteForceCandidates,MeanLowerSamples,StdLowerSamples,MeanLowerSamplingFraction,StdLowerSamplingFraction,MinLowerSamplesPerCellMatrix,MaxLowerSamplesPerCellMatrix,MeanLowerHoffman,StdLowerHoffman,MeanLowerSeconds,StdLowerSeconds,MeanBruteForceHoffman,StdBruteForceHoffman,MeanBruteForceSeconds,StdBruteForceSeconds,MeanPVZHoffman,StdPVZHoffman,MeanPVZSeconds,StdPVZSeconds,MeanLowerAbsoluteError,StdLowerAbsoluteError,MeanPVZAbsoluteError,StdPVZAbsoluteError,MeanPVZSpeedup,StdPVZSpeedup,MeanUpperHoffman,StdUpperHoffman,MeanUpperSeconds,StdUpperSeconds,MeanUpperAbsoluteError,StdUpperAbsoluteError
function_H_pq,2,3,6,all_zero,ceil_one_tenth_brute_force_candidates_per_cell_matrix,30,6.0,0.0,42.0,0.0,6.0,0.0,0.14285714285714282,2.823006427041851e-17,1,1,0.6781095589571406,0.46051955591974747,0.1339372132333333,0.028022494047512735,1.5299661116595118,0.5382337158808744,0.3919698738333333,0.11491559976733827,1.5299661116595118,0.5382337158808744,0.20731824923333325,0.040769019806321,0.8518565527023708,0.5836747673733469,0.0,0.0,1.9264439127500614,0.588996062235836,4.366149182627799,1.7012195743435155,0.08658706076666665,0.02228418719513645,2.8361830709682887,1.2332752942190142
function_H_pq,3,4,9,all_zero,ceil_one_tenth_brute_force_candidates_per_cell_matrix,30,12.0,0.0,372.0,0.0,48.0,0.0,0.12903225806451604,8.469019281125553e-17,4,4,0.8642366421136318,0.20900414962288358,0.5370003236999998,0.239265656813462,1.4312413968981257,0.372048262304296,2.6029499128333327,0.34480704769203857,1.4312413968981257,0.372048262304296,0.3682525100666666,0.07625068906844441,0.5670047547844936,0.34611763862759354,1.258252761241844e-16,2.5204808693737353e-16,7.324434148554463,1.7156771456093287,7.212581044944329,2.59286729253534,0.1649938556,0.07382605447982106,5.781339648046207,2.391735022757237
function_H_pq,5,4,8,all_zero,ceil_one_tenth_brute_force_candidates_per_cell_matrix,30,20.0,0.0,2540.0,0.0,260.0,0.0,0.1023622047244094,4.2345096405627763e-17,13,13,1.9773804238802022,1.030005659157603,2.095431208533333,0.23471966246428633,2.8625631945777976,1.0283539115469467,14.647767762299997,1.2485003414185756,2.8625631945777967,1.0283539115469462,0.5491386226666667,0.1891121831134553,0.8851827706975947,0.7210185385095159,6.217248937900876e-16,9.659358343985362e-16,28.50811311709448,7.038379721355556,30.791219485956034,47.69303741580716,0.2805792210333333,0.06385935162465366,27.92865629137824,47.531078440332735
function_H_pq,7,3,12,all_zero,ceil_one_tenth_brute_force_candidates_per_cell_matrix,30,21.0,0.0,5355.0,0.0,546.0,0.0,0.10196078431372554,5.646012854083702e-17,26,26,1.0622345646707074,0.3000753727747113,4.236204666633334,0.6874767792593308,1.2494389887317532,0.3982388705403726,29.89043901786666,1.351795590900967,1.2494389887317532,0.3982388705403726,0.5479956302333334,0.06349666458707441,0.18720442406104565,0.2612386070020078,2.516505522483688e-16,3.9326234720845927e-16,55.393397433772705,8.07510511412163,10.188069145478357,4.368332196610521,0.4662846948666667,0.2452374444405863,8.938630156746601,4.094220078602685
function_H_pq,7,8,2,all_zero,ceil_one_tenth_brute_force_candidates_per_cell_matrix,30,10.366666666666667,1.2994251602637368,943.3666666666667,118.24768958400001,103.66666666666667,12.994251602637364,0.10989010989010993,4.2345096405627763e-17,10,10,209.23455284452572,490.0934962373427,1.0532120516333334,0.17869872900891814,923.18148857125,3283.9782315186317,5.9983208472,0.736420732945517,923.1814885713273,3283.9782315190064,0.4537793604666666,0.09166168497379826,713.946935726725,3285.458858353729,7.790378712494809e-11,3.778144010613105e-10,13.578665078592078,2.5135053819620032,21255.867617199874,76366.46303926005,0.26398554520000006,0.04165195927118299,20332.68612862862,76483.94986357469
function_H_pq,6,9,2,all_zero,ceil_one_tenth_brute_force_candidates_per_cell_matrix,30,10.433333333333334,1.2228664272317618,949.4333333333333,111.28084487809035,104.33333333333333,12.228664272317623,0.10989010989010993,4.2345096405627763e-17,10,10,3463.7154670984937,17531.4616299212,1.1083817295333336,0.19355131124116357,4060.0969871651205,17547.157647101223,6.226841257133332,1.0009323145305498,4060.0969871842167,17547.1576472051,0.48963163753333344,0.13186516180231367,596.3815201051896,2183.803649965235,1.915140591298344e-8,1.0471456982270949e-7,13.227619038305106,2.8712601924113987,12800.793231141968,22033.210923328817,0.2913203945666666,0.05861905256498911,8740.696243976847,13449.177504969743
\end{filecontents*}
\pgfplotstabletypeset[
    col sep=comma,
    columns={MP,MQ,N,
        MeanLowerAbsoluteError,StdLowerAbsoluteError,
        MeanLowerSeconds,StdLowerSeconds,
        MeanBruteForceSeconds,StdBruteForceSeconds,
        MeanPVZAbsoluteError,StdPVZAbsoluteError,
        MeanPVZSeconds,StdPVZSeconds,
        MeanUpperAbsoluteError,StdUpperAbsoluteError,
        MeanUpperSeconds,StdUpperSeconds},
    every head row/.style={
        before row={
            \toprule
            $s$ & $t$ & $n$
            & \multicolumn{2}{c}{$H_L$ Error}
            & \multicolumn{2}{c}{$H_L$ Time}
            & \multicolumn{2}{c}{$H_{\mathrm{BF}}$ Time}
            & \multicolumn{2}{c}{$H_{\mathrm{PVZ}}$ Error}
            & \multicolumn{2}{c}{$H_{\mathrm{PVZ}}$ Time}
            & \multicolumn{2}{c}{$H^U$ Error}
            & \multicolumn{2}{c}{$H^U$ Time}\\
        },
        after row=\midrule,
        output empty row
    },
    every last row/.style={after row=\bottomrule},
    every column/.style={column type=r, fixed, fixed zerofill, precision=3},
    columns/MP/.style={fixed, precision=0},
    columns/MQ/.style={fixed, precision=0},
    columns/N/.style={fixed, precision=0},
    columns/MeanLowerAbsoluteError/.style={sci, sci zerofill, precision=2},
    columns/StdLowerAbsoluteError/.style={column type={@{\,$\pm$\,}r}, sci, sci zerofill, precision=2},
    columns/StdLowerSeconds/.style={column type={@{\,$\pm$\,}r}},
    columns/StdBruteForceSeconds/.style={column type={@{\,$\pm$\,}r}},
    columns/MeanPVZAbsoluteError/.style={sci, sci zerofill, precision=2},
    columns/StdPVZAbsoluteError/.style={column type={@{\,$\pm$\,}r}, sci, sci zerofill, precision=2},
    columns/StdPVZSeconds/.style={column type={@{\,$\pm$\,}r}},
    columns/MeanUpperAbsoluteError/.style={sci, sci zerofill, precision=2},
    columns/StdUpperAbsoluteError/.style={column type={@{\,$\pm$\,}r}, sci, sci zerofill, precision=2},
    columns/StdUpperSeconds/.style={column type={@{\,$\pm$\,}r}}
]{data/hoffman-signomials/hoffman_summary.csv}%
}
\end{table}

The results show a clear trade-off between computational cost and the quality of the resulting estimate. The PVZ algorithm recovers the same Hoffman constant as brute-force enumeration in all configurations (up to a small error), as expected (and so provides a numerical verification of our implementation). The very small nonzero errors reported for PVZ in some configurations are for numerical reasons. The PVZ algorithm also substantially reduces computation time for the larger problems. We see that the brute-force runtime increases substantially across the larger configurations, while the PVZ computation time remains comparatively small and stable. This shows the advantage of the PVZ algorithm and of avoiding exhaustive evaluation of candidate subsets, which is required by the brute-force method.

The randomized lower bound is also inexpensive to compute and remains close to the exact value in these examples, despite studying only a fraction of the candidate subsets. Very small errors may appear as zero at the displayed
precision. The upper bound is by far the cheapest quantity to compute, but is considerably looser. This behavior is consistent with the motivation: rather than accurately estimating the Hoffman constant, the upper bound provides a computationally inexpensive and conservative quantity that can still be used to obtain a guaranteed search domain.

\begin{figure}[H]
    \centering
    \begin{subfigure}[b]{0.48\textwidth}
        \centering
        \includegraphics[width=0.8\textwidth]{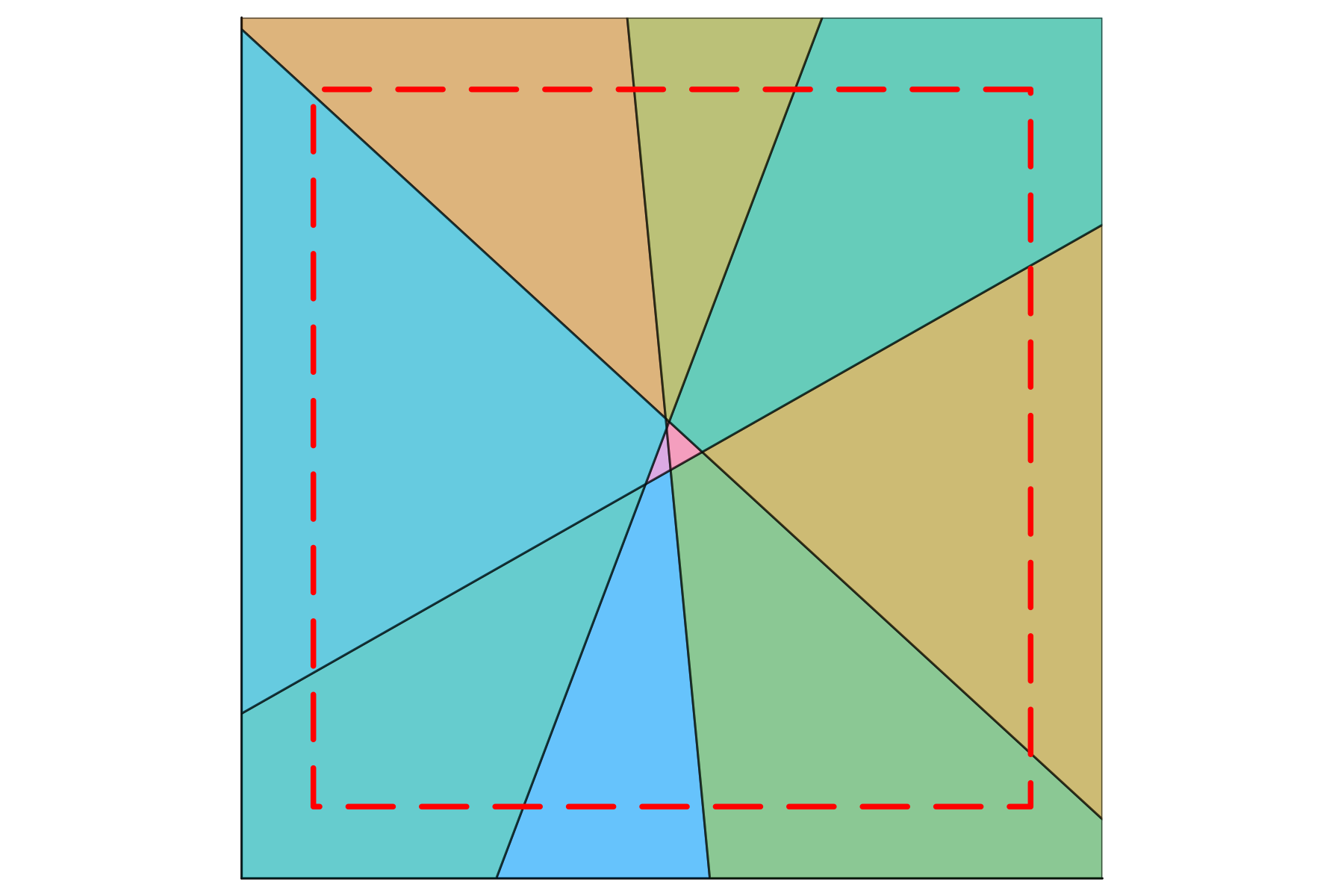}
        \caption*{Exact}
    \end{subfigure}
    \hfill
    \begin{subfigure}[b]{0.48\textwidth}
        \centering
        \includegraphics[width=0.8\textwidth]{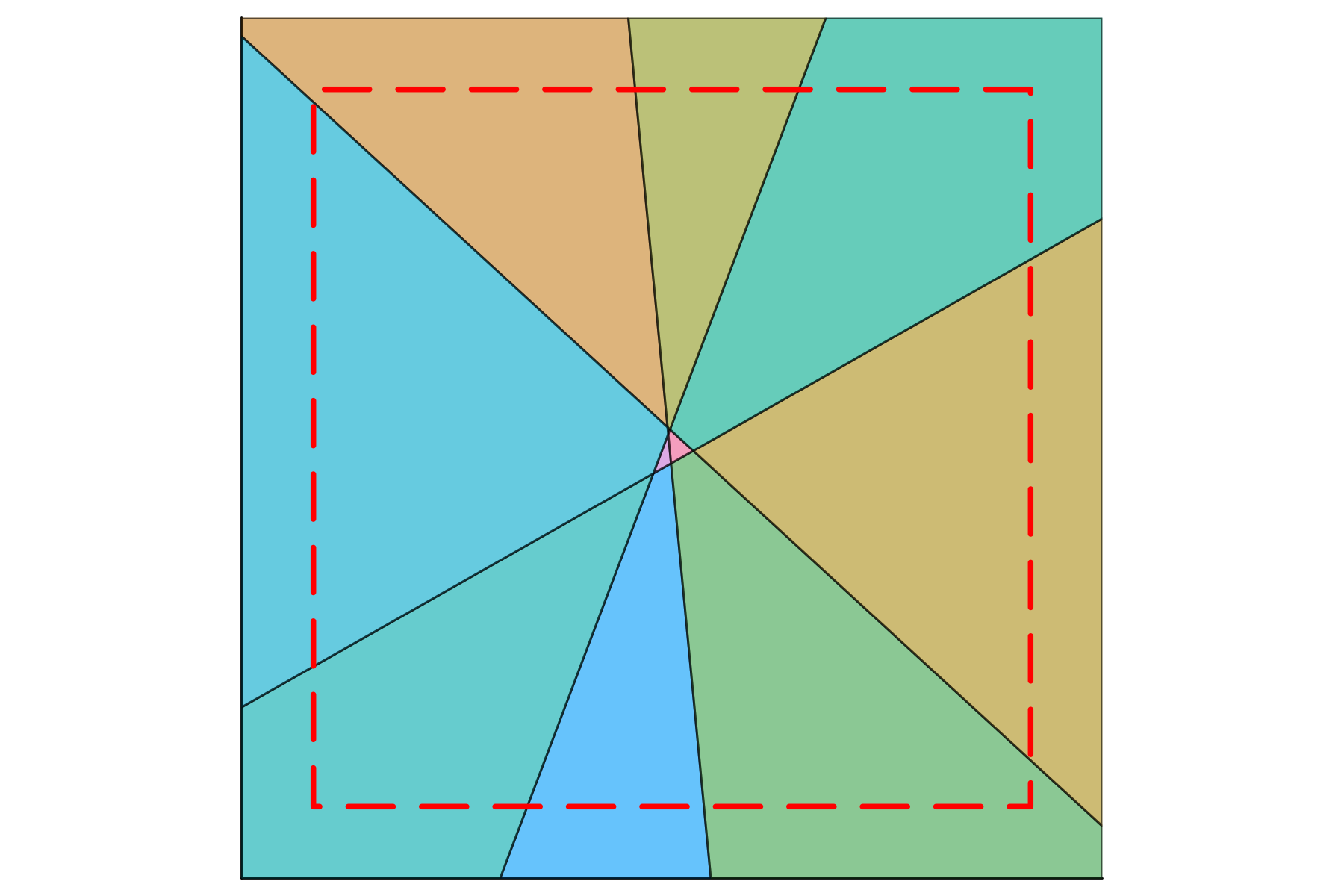}
        \caption*{Upper-bound estimate.}
    \end{subfigure}
    \caption{Effective search domains for a randomly initialized one-hidden-layer neural network of width 4. We compute the radius bound using the exact Hoffman constant (left) and an upper bound on the Hoffman constant (right). Both boxes are guaranteed to intersect every linear region of the network; the upper-bound estimate yields a more conservative search domain. The panels are displayed at different scales.
    }
    \label{fig:effective_radius}
\end{figure}

\paragraph{Constructing an Effective Search Domain.} We next demonstrate how the Hoffman computations can be used in practice for their intended geometric goal. For a randomly initialized one-hidden-layer neural network of width 4, we compute the effective-radius bound from \cref{prop:effective-radius} using both the exact Hoffman constant obtained by the PVZ algorithm and the upper bound from \cref{alg:upper_hoffman}. The resulting search domains are shown in \cref{fig:effective_radius}.

In both cases, the resulting box intersects every linear region of the network. Using the exact Hoffman constant results in a tighter search domain, whereas replacing it by the computationally cheaper upper bound yields a more conservative domain. The latter nevertheless retains the essential guarantee: all linear regions can be intersected within a finite, explicitly computable subset of the input space.

Thus, the Hoffman-based construction turns the otherwise unbounded problem of searching for the linear regions of a neural network into a bounded one with an explicit geometric guarantee.

%% file: arXiv/discussion_new.tex
\section{Discussion}

In this work, we developed a computational tropical geometry framework for the symbolic analysis of neural networks. Building on the representation of piecewise-linear neural networks as tropical rational maps, we showed how their linear regions can be computed exactly as unions of polyhedra and introduced complementary algebraic and geometric quantities for studying these representations. In particular, the Hoffman constant provides quantitative control over the location of linear regions in the input space, while non-redundant monomials capture the algebraic complexity of tropical representations. These methods are implemented in \textsc{TropicalNN.jl}, which provides an extensible computational framework for manipulating and analyzing neural networks as tropical algebraic objects.

Further to exact enumeration, the geometric information provided by our framework suggests several potential applications. The distribution and concentration of linear regions throughout the input space may provide information about structures learned by a network and how these evolve during training. More specifically, our Hoffman-based bounds address a practical challenge in exact linear-region counting, which is that several existing methods operate on bounded input domains \citep[e.g.,][]{humayun2023splinecam, serra2020empirical}. Our results provide an explicitly computable bounded domain guaranteed to intersect every linear region, which allows a global coverage guarantee for such bounded-domain analyses.

\paragraph{Limitations.} The main limitation of exact symbolic analysis is computational. Although we provide correctness guarantees for the exact algorithms that we developed, they ultimately require combinatorial enumeration and the solution of potentially large collections of linear programs. This illustrates the intrinsic difficulty of the underlying problems rather than only limitations of the implementation. Computing the global Hoffman constant---which is related to the Stewart--Todd condition measure of a matrix---is $\mathsf{NP}$-hard in general \citep{pena2024easily}, while determining whether a deep ReLU network has more than $K$ linear regions is also $\mathsf{NP}$-hard \citep{wang2022estimation}. Therefore, the exact methods developed here are most naturally suited to settings where symbolic and geometric information is of particular theoretical interest, rather than to large-scale enumeration for arbitrary modern network architectures. The Hoffman-constant approximation methods and the pruning procedures developed in this work provide practical approaches for reducing some of these computational costs.

\paragraph{Future Directions.} Our framework opens several directions for further research at the interface of geometry, algebra, and computation. On the geometric side, the polyhedral subdivisions arising from tropical representations of neural networks possess rich structure beyond their number of regions. Other geometric quantities such as region volumes, boundedness, face structure, and other combinatorial properties may reveal how network architecture and training shape the geometry of the represented function; related questions have begun to be explored in \cite{liu2023relu,gaines2026characterizing}. The symbolic and polyhedral functionality of \textsc{TropicalNN.jl} provides tools for
investigating such questions systematically.

On the algebraic side, the number of non-redundant monomials provides a natural measure of the complexity of a particular tropical representation, but is not in general an invariant of the represented function. A natural next step is therefore to study minimal tropical
representations, extending the theoretical direction of \cite{tran2024minimal}, with the goal of defining representation-independent notions of algebraic complexity for neural networks.

Finally, algorithmic improvement is an important direction for future work. The computations developed here often require solving families of closely related linear programs arising from the same underlying polyhedral subdivision. 
Specialized optimization methods that reuse information across these problems could substantially improve computational efficiency and extend the range of networks which may be studied using exact symbolic analysis.